\documentclass[a4paper, 10pt, conference]{ieeeconf} 
\usepackage{FG2023}

\FGfinalcopy 

\usepackage{xcolor,soul}

\usepackage[breaklinks=true,bookmarks=false]{hyperref}

\IEEEoverridecommandlockouts                              
\title{\LARGE \bf
StyleMask: Disentangling the Style Space of StyleGAN2 for Neural Face Reenactment
}

\author{\parbox{16cm}{\centering
    {\large Stella Bounareli$^1$,  Christos Tzelepis$^2$,  Vasileios Argyriou$^1$,  Ioannis Patras$^2$, Georgios Tzimiropoulos$^2$ }\\
    {\normalsize
    $^1$ School of Computer Science and Mathematics, Kingston University London\\
    $^2$ School of Electronic Engineering and Computer Science, Queen Mary University of London}}
    \thanks{This work was supported by EU H2020 project AI4Media No. 951911.}
}

\usepackage{cite}
\usepackage{amsmath,amssymb,amsfonts}
\usepackage{algorithmic}
\usepackage{graphicx}
\usepackage{textcomp}
\usepackage{xcolor}
\usepackage{subcaption}
\usepackage{color,colortbl}
\usepackage{xcolor,soul}
\definecolor{LightCyan}{rgb}{0.88,1,1}
\definecolor{Gray}{gray}{0.9}

\begin{document}

\ifFGfinal
\thispagestyle{empty}
\pagestyle{empty}
\else
\author{Anonymous FG2023 submission\\ Paper ID \FGPaperID \\}
\pagestyle{plain}
\fi

\maketitle

\begin{abstract}
In this paper we address the problem of neural face reenactment, where, given a pair of a source and a target facial image, we need to transfer the target's pose (defined as the head pose and its facial expressions) to the source image, by preserving at the same time the source's identity characteristics (e.g., facial shape, hair style, etc), even in the challenging case where the source and the target faces belong to different identities. In doing so, we address some of the limitations of the state-of-the-art works, namely, a) that they depend on paired training data (i.e., source and target faces have the same identity), b) that they rely on labeled data during inference, and c) that they do not preserve identity in large head pose changes. More specifically, we propose a framework that, using unpaired randomly generated facial images, learns to disentangle the identity characteristics of the face from its pose by incorporating the recently introduced \textit{style space} $\mathcal{S}$~\cite{wu2021stylespace} of StyleGAN2~\cite{karras2020analyzing}, a latent representation space that exhibits remarkable disentanglement properties. By capitalizing on this, we learn to successfully mix a pair of source and target style codes using supervision from a 3D model. The resulting latent code, that is subsequently used for reenactment, consists of latent units corresponding to the facial pose of the target only and of units corresponding to the identity of the source only, leading to notable improvement in the reenactment performance compared to recent state-of-the-art methods. In comparison to state of the art, we quantitatively and qualitatively show that the proposed method produces higher quality results even on extreme pose variations. Finally, we report results on real images by first embedding them on the latent space of the pretrained generator. We make the code and the pretrained models publicly available at: \url{https://github.com/StelaBou/StyleMask}.
\end{abstract}

\section{Introduction}\label{sec:introduction}
    Generative Adversarial Networks (GANs)~\cite{goodfellow2014generative} have emerged as the leading generative paradigm improving image synthesis to levels of exceptional realism. State-of-the-art GANs~\cite{karras2019style,karras2020analyzing} are able to synthesize high quality fake images that are indistinguishable from real ones. Leveraging the photo-realistic image generation ability of GANs, many researchers focus on areas such as super resolution~\cite{wang2018esrgan}, image editing~\cite{shen2020interfacegan,voynov2020unsupervised,tzelepis2021warpedganspace,oldfield2021tensor,oldfield2022panda}, real image inversion~\cite{tov2021designing,alaluf2021restyle,alaluf2021hyperstyle}, and neural face reenactment~\cite{zakharov2020fast,bounareli2022finding,ren2021pirenderer}.
    
    \begin{figure}
        \centering
        \includegraphics[width=1.0\linewidth]{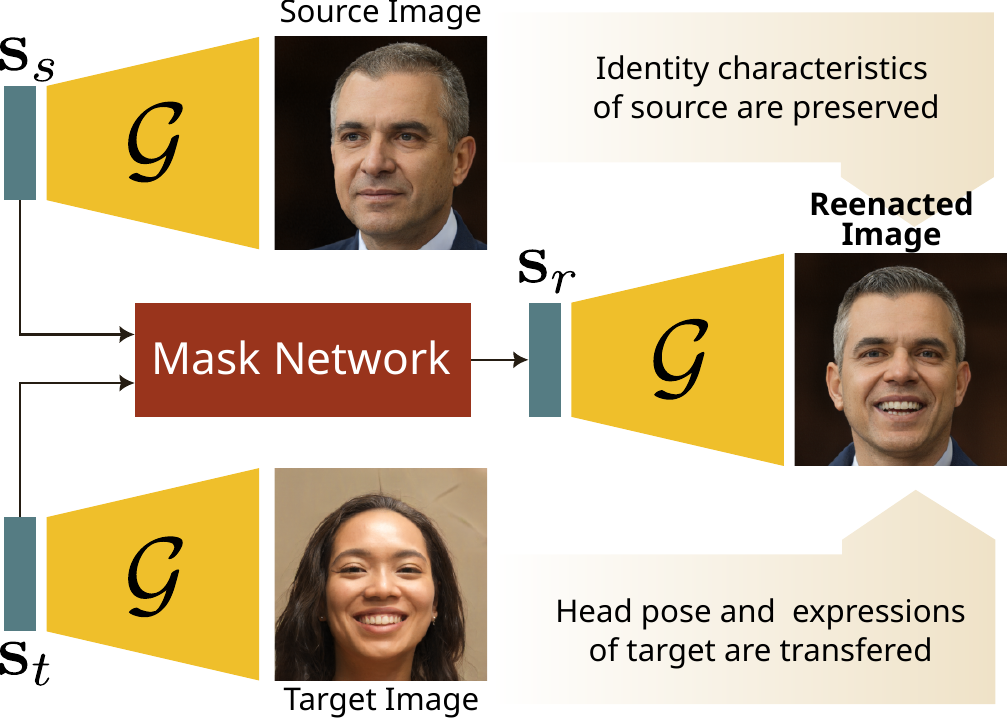}
        \caption{Overview of the proposed \textit{StyleMask}. We perform face reenactment by learning to disentangle in the style space $\mathcal{S}$~\cite{wu2021stylespace} of a pre-trained StyleGAN2~\cite{karras2020analyzing} $\mathcal{G}$ the identity of a source image (style code $\mathbf{s}_s$) and the head pose and expression of a target image (style code $\mathbf{s}_t$), leading to a reenacted image (style code $\mathbf{s}_r$) capturing the identity of the source and the head pose and expression of the target.}
        \label{fig:architecture_small}
    \end{figure}
    
    Despite the unprecedented photo-realism of the generated faces given by state-of-the-art GANs~\cite{karras2019style,karras2020analyzing}, the controllability of generation remains an open and challenging problem. In order to understand the structure of the latent space and the underlying generative factors of GANs, the research community has recently directed its efforts towards discovering interpretable/disentangled directions in the latent space of pre-trained generators~\cite{shen2020interfacegan,2020ganspace,voynov2020unsupervised,tzelepis2021warpedganspace, oldfield2021tensor}; that is, latent directions travelling across which gives rise to generations where only a single (or a very few) generative factors are activated. Typically, these latent directions are used for image editing (e.g., the editing of facial images in terms of certain attributes, head pose, facial expressions, etc). However, explicit controllability of such methods is very limited and their evaluation relies either on laborious manual annotation~\cite{voynov2020unsupervised, 2020ganspace} or on certain pre-trained detectors~\cite{tzelepis2021warpedganspace}.
    
    Neural face reenactment is the task where, given a pair of a source and a target image, it is required to transfer the target's pose (i.e., the head pose and its facial expressions) to the source image, by preserving at the same time the source's identity characteristics (e.g., facial shape, hair style, etc). This is a challenging problem where controllability and disentanglement in the GAN generation process is of utmost importance. A few recent works propose to address the problem by incorporating pre-trained GANs~\cite{bounareli2022finding,nitzan2020face}, or by training conditional generative models~\cite{zakharov2020fast,doukas2020headgan,ren2021pirenderer}. The core challenge of the face reenactment task consists in disentangling the identity characteristics from the head pose and expression. This is typically tackled by the research community~\cite{siarohin2019first,zakharov2020fast,ren2021pirenderer,doukas2020headgan} by trying to learn disentangled representations using supervised training on paired data (i.e., images where the source and the target faces have the same identity) on large video datasets, such as VoxCeleb~\cite{Nagrani17,Chung18b}. However, this poses a limitation to the applicability of such methods on cross-subject reenactment, where the source and target faces have different identities~\cite{zakharov2019few}. By contrast, our method relies solely on randomly generated images (using a pre-trained StyleGAN2) in learning to disentangle the identity characteristics from the head pose and expression.
    
    In this work, we propose to learn to disentangle the identity information of the source image from the facial pose (head pose and expression) of the target image by learning to mask and mix the corresponding channels in the style space $\mathcal{S}$~\cite{wu2021stylespace} of a pre-trained StyleGAN2~\cite{karras2020analyzing} without using paired training data (Fig.~\ref{fig:architecture_small}). Inspired by the approach of StyleFusion~\cite{kafri2021stylefusion}, we utilize the remarkable disentanglement properties of the recently proposed style space $\mathcal{S}$ and we learn how to mask and mix the source and the target style codes in order to arrive at a style code that faithfully reconstructs the source identity and effectively transfers the target head pose and facial expressions. For imposing identity preservation and head pose transfer we use additional supervision by pre-trained networks~\cite{deng2019arcface,feng2021deca}, and we further mitigate any visual artifacts produced when the source and target faces have large head pose differences by adding a recurrent cycle consistency objective. By contrast to~\cite{kafri2021stylefusion}, instead of training a separate network for each region of interest (e.g., for hair and eyes) using supervision from segmented masks, we train a \textit{single} network using supervision from a 3D shape model~\cite{feng2021deca}. Moreover, our model is trained using randomly generated images, instead of using paired ones (where the source and the target faces have the same identity), allowing the effective reenactment between faces of different identities. The main contributions of this paper can be summarized as follows:
    \begin{enumerate} 
        \item We propose a method for learning to disentangle the identity and the pose (i.e., head pose and facial expressions) of human faces using the style space $\mathcal{S}$~\cite{wu2021stylespace} of StyleGAN2~\cite{karras2020analyzing} for the task of face reenactment.
        \item Our method is trained on randomly generated synthetic images, instead of depending on paired training data, allowing this way the effective reenactment between faces of different identities. Moreover, our method uses supervision from pre-trained models only and we show that it can be straightforwardly adjusted on real images capitalizing on real image inversion methods~\cite{tov2021designing,alaluf2021hyperstyle}.
        \item We leverage a recurrent cycle consistency objective to mitigate visual artifacts produced when the source and target faces have large head pose differences.
        \item We train our model using StyleGAN2~\cite{karras2020analyzing} pre-trained on the FFHQ dataset~\cite{karras2019style} and compare with the following state-of-the-art methods: a) StyleFusion~\cite{kafri2021stylefusion}, ID-disentanglement~\cite{nitzan2020face}, and StyleFlow~\cite{abdal2021styleflow} that use pre-trained GANs, and b) Fast Bi-layer~\cite{zakharov2020fast} and PIR~\cite{ren2021pirenderer} that perform face reenactment by training from scratch controllable generative models using large video datasets~\cite{Nagrani17,Chung18b}. We show, both qualitatively and quantitatively, that in comparison to state of the art, our method disentangles better the identity and the facial pose, leading to more faithful identity preservation and effective head pose and expression transfer.
    \end{enumerate}

\section{Related Work}\label{sec:related_work}
    
    StyleGAN2~\cite{karras2020analyzing} has emerged as the dominating architecture for image synthesis due to its remarkable ability to generate photo-realistic fake images, usually indistinguishable from the real ones. In order to understand the underlying generative factors of GANs, the research community focuses on discovering interpretable and controllable directions in the latent space of pre-trained generators~\cite{shen2020interfacegan,2020ganspace,voynov2020unsupervised,tewari2020stylerig,kafri2021stylefusion,tzelepis2021warpedganspace, oldfield2021tensor}. Typically, these latent directions are used for image editing (e.g., the editing of facial images in terms of certain attributes, head pose, facial expressions, etc). Voynov and Babenko~\cite{voynov2020unsupervised}, introduced an unsupervised and model-agnostic method for discovering interpretable linear paths in the latent space of pre-trained GANs. Tzelepis et al.~\cite{tzelepis2021warpedganspace} built upon~\cite{voynov2020unsupervised} in order to discover non-linear interpretable latent paths. Both~\cite{voynov2020unsupervised} and~\cite{tzelepis2021warpedganspace} optimize latent paths so as the induced image transformations are easily distinguishable by a discriminator network. Moreover, GANSpace~\cite{2020ganspace} performs Principal Components Analysis (PCA) on deep features at the early layers of the generator and finds directions in the latent space that best map to those deep PCA vectors, arriving at a set of non-orthogonal directions in the latent space. Similarly, in~\cite{shen2021closedform}, the authors propose to perform eigenvector decomposition to discover the most meaningful directions. All the aforementioned unsupervised methods are able to discover directions on the latent space of GANs that correspond to meaningful transformations on the generated images. However, in contrast to our method, explicit controllability of such methods is very limited, introducing severe changes in the identity characteristics, and their evaluation relies either on laborious manual annotation~\cite{voynov2020unsupervised, 2020ganspace} or on certain pre-trained detectors~\cite{tzelepis2021warpedganspace}. 
    
    In a similar line of research, supervised methods rely on external supervision either from pretrained attribute classifiers (e.g.,~\cite{shen2020interfacegan}) or vision-language models (e.g.,~\cite{patashnik2021styleclip, tzelepis2022contraclip}), in order to assign labels on the generated images and discover the disentangled directions that control the labeled attributes. In~\cite{tewari2020stylerig}, Tewari et al. introduce StyleRig that combines the 3D Morphable Models (3DMM)~\cite{blanz1999morphable} and StyleGAN2~\cite{karras2020analyzing} to control some facial attributes (e.g., head pose, smile, etc) on the generated images. Despite its success in controlling those attributes, editing cannot be conducted simultaneously for more than one attribute. This introduces a crucial limitation, rendering it inapplicable in the problem of face reenactment. In order to disentangle the identity characteristics from other facial attributes (such as head pose and expression), Nitzan et al.~\cite{nitzan2020face} propose a framework with two encoders, where they first learn a disentangled representation that combines the identity and attribute features extracted from the corresponding encoders and then map this representation into the $\mathcal{W}$ latent space of a pre-trained StyleGAN. Their method is able to disentangle the identity from the facial pose, however the reenacted images do not faithfully preserve the identity characteristics of the source face. Finally, in StyleFlow~\cite{abdal2021styleflow}, the authors introduce a method of finding non-linear paths in the latent space of StyleGAN by learning conditional continuous normalizing flows using supervision from multiple attribute classifiers and regressors. By contrast to StyleFlow~\cite{abdal2021styleflow}, which requires labels both during training and inference, our method requires only an unlabeled pair of a source and a target image during inference.
    
    \begin{figure*}[t!]
        \centering
        \includegraphics[width=0.85\textwidth]{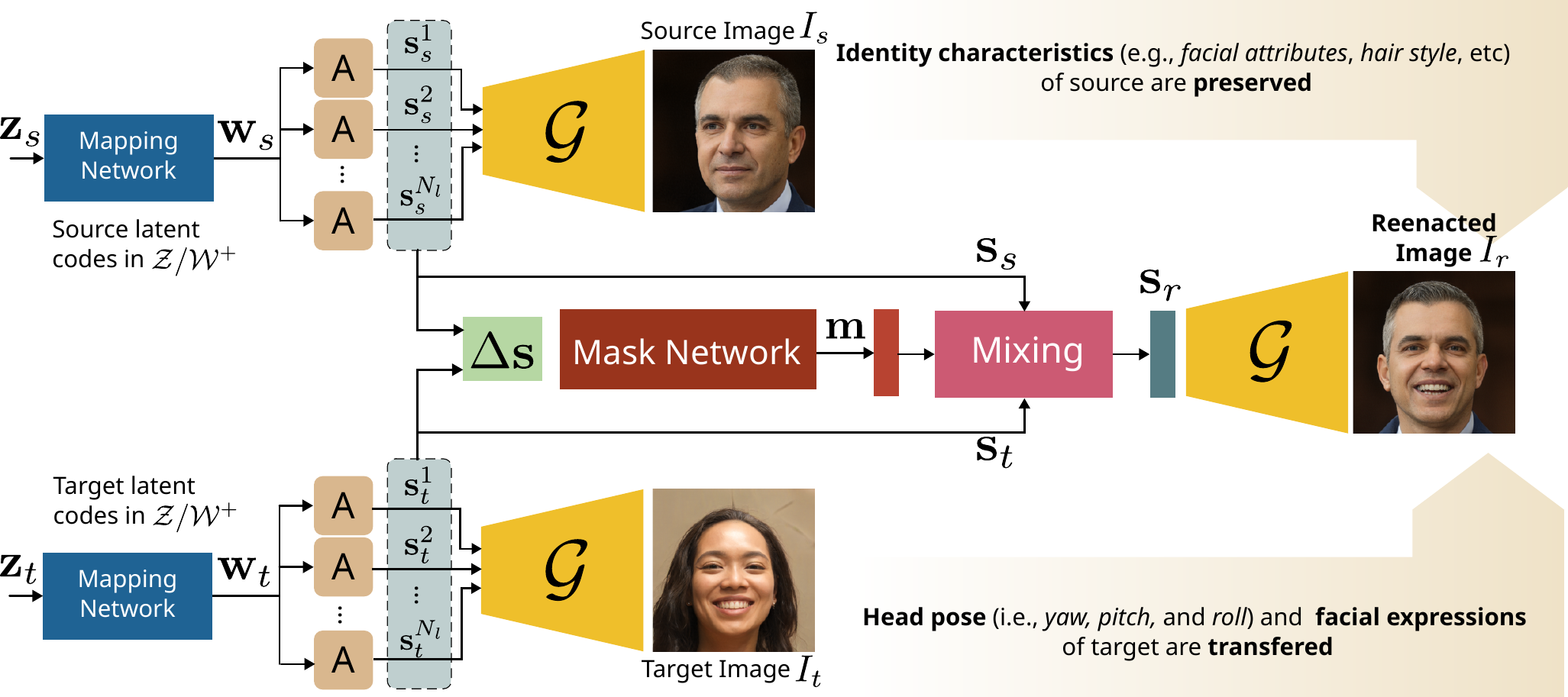}
        \caption{StyleMask -- proposed framework: Given a pair of a source ($\mathbf{s}_s$) and a target ($\mathbf{s}_t$) style codes, we learn a mask vector $\mathbf{m}$ such that the reenacted style code $\mathbf{s}_r$ consists of the style channels of the target style code that correspond to the facial pose (head pose and expression) and the style channels of the source style code that correspond to the identity characteristics. The Mask Network acts on the difference $\Delta\mathbf{s}=\mathbf{s}_s-\mathbf{s}_t$ and outputs the mask vector $\mathbf{m}$, which then is used to calculate the reenacted style code using Eq.~\ref{eq:mask_eq}.}
        \label{fig:architecture}
    \end{figure*}
    
    A method which is closely related to ours is StyleFusion~\cite{kafri2021stylefusion}. In~\cite{kafri2021stylefusion}, Kafri et al. propose to learn to disentangle and control several attributes of the generated images based on discovered segmented semantic regions of facial images using a hierarchical architecture. More specifically, they learn to disentangle a pair of different image regions (e.g., hair and eyes) on each level, using a multi-step training process and three different style codes. That is, in the first step of their training stage, they optimize a network to align the two style codes into the spatial location (i.e., head pose) of the third style code. Then, on the second step, given the aligned style codes, they train a new network that learns to disentangle the desired semantic regions and create a harmonised image that depicts the spatial location (i.e., head pose) of the third style code and the semantic regions (i.e., hair and eyes) of the aligned style codes. By contrast, we train a \textit{single} network that learns to generate a harmonized style code by disentangling the identity attributes (i.e., hair style/color and facial shape) and the semantic attributes (i.e., head pose and expression) at once. Finally, StyleFusion~\cite{kafri2021stylefusion} focuses on semantic image editing and it proposes to disentangle the different image regions by training different networks for each region of interest in a sequential manner, using supervision from segmented masks. Capitalizing on their idea, we focus on the face reenactment task where the goal is to disentangle the identity characteristics from the facial pose. We achieve this by training our network using supervision from a 3D shape model~\cite{feng2021deca}.
    
    Neural face reenactment is a challenging task, notably more complex than image editing, since it requires the simultaneous transfer of the head pose and the facial expressions of a target face into a source face by preserving at the same time the identity characteristics of the source face. Most of the state-of-the-art face reenactment methods~\cite{zakharov2019few,zakharov2020fast,meshry2021learned,wang2021one,doukas2020headgan,ren2021pirenderer} rely on paired data (i.e., source and target faces having the same identity) in order to train controllable generative models. Another line of works is based on incorporating facial landmarks~\cite{zakharov2019few,zakharov2020fast,meshry2021learned} -- however, facial landmarks contain important information about the identity of the source face (i.e., facial shape), which impedes the applicability of such methods in cross-subject face reenactment~\cite{zakharov2019few,zakharov2020fast}. Recently, the works of~\cite{ren2021pirenderer} and~\cite{doukas2020headgan} propose the use of 3D shape models to transfer the facial motion, and in~\cite{bounareli2022finding}, Bounareli et al. introduce a novel approach on neural face reenactment by finding directions in the latent space of a pretrained StyleGAN2 on the VoxCeleb dataset~\cite{Nagrani17}, that control the facial attributes (i.e., head pose and expression) using a 3D shape model~\cite{feng2021deca}. Similarly, during training we also use a 3D shape model to disentangle the facial pose from the identity, however during inference our model relies solely on a source and a target style code.

\section{Method}\label{sec:method}

    In this Section, we will briefly discuss the different latent spaces of StyleGAN2~\cite{karras2020analyzing} in Section~\ref{ssec:latent_spaces} and then we will present our method for face reenactment in detail in Section~\ref{ssec:method_S}. An overview of the proposed framework is given in Fig.~\ref{fig:architecture_small}. In a nutshell, we perform neural face reenactment using the style space $\mathcal{S}$~\cite{wu2021stylespace} of StyleGAN2~\cite{karras2020analyzing}, where, given a pair of style codes $\mathbf{s}_s$ and $\mathbf{s}_t$, corresponding to the source and the target face images, respectively, we learn to disentangle the identity and the facial pose (head pose and expression) channels in $\mathcal{S}$. Our intuition to do so is driven by the fact that the style space $\mathcal{S}$ provides a remarkable disentangled structure where changing a single channel leads to change of a single semantic attribute~\cite{wu2021stylespace}. More specifically, we optimize a \textit{Mask Network} that takes as input the style codes $\mathbf{s}_s$ and $\mathbf{s}_t$ and outputs a new reenacted style code $\mathbf{s}_r$. Then, $\mathbf{s}_r$ is passed through the generator $\mathcal{G}$ and generates an image that has the identity of the source image and the facial pose of the target image. A detailed overview of the proposed framework is given in Fig.~\ref{fig:architecture}.
    
    \subsection{Latent spaces of StyleGAN2}\label{ssec:latent_spaces}
        As discussed in previous sections, synthetic image editing in GANs is typically addressed by the research community by manipulating latent representations on a given GAN latent space (e.g.,~\cite{voynov2020unsupervised,wu2021stylespace,patashnik2021styleclip}). StyleGAN2~\cite{karras2020analyzing} provides a basic latent space $\mathcal{Z}\subset\mathbb{R}^{512}$ where samples are drawn from the multi-variate standard Gaussian $\mathcal{N}(\mathbf{0},\mathbf{I})$, before they are passed through a Multilayer Perceptron (MLP) to produce the intermediate latent representations $\mathbf{w}\in\mathcal{W}\subset\mathbb{R}^{512}$. Then, $\mathbf{w}$ latent codes are fed into the $N_l$ layers of the synthesis network. $\mathcal{W}$ space has proven to be more disentangled than $\mathcal{Z}$ space~\cite{karras2020analyzing}, rendering it the most common choice for synthetic image manipulation (e.g.,~\cite{voynov2020unsupervised,tzelepis2021warpedganspace,2020ganspace}). Moreover, Abdal et al.~\cite{abdal2019image2stylegan} propose to extend $\mathcal{W}$ into the $\mathcal{W}^+\subset\mathbb{R}^{N_l\times512}$ latent space, where each layer of the synthesis network takes as input a different latent code $\mathbf{w^+}$. This space is commonly used on real image inversion methods (e.g.,~\cite{tov2021designing,alaluf2021restyle}), since it provides greater expressiveness than the $\mathcal{W}$ space. 
        
        Recently, Wu et al.~\cite{wu2021stylespace} introduce a new latent space for StyleGAN2, termed as the style space $\mathcal{S}$. In StyleGAN2, each latent code $\mathbf{w}\in\mathcal{W}$ is transformed using affine transformation $A$ into channel-wise style vectors $\mathbf{s}\in\mathcal{S}$, which are then passed into the different layers of the generator (this is illustrated in the left part of Fig.~\ref{fig:architecture}). In~\cite{wu2021stylespace}, the authors show that this space is more disentangled than the $\mathcal{Z}$ and $\mathcal{W}$/$\mathcal{W}^+$ spaces and, as a result, it enables better semantic editing of synthetic images. We note that a StyleGAN2 model with $1024\times1024$ image resolution has $N_l=18$ layers, $\mathcal{Z},\mathcal{W}\subset\mathbb{R}^{512}$, $\mathcal{W}^+\subset\mathbb{R}^{18\times512}$ and $\mathcal{S}\subset\mathbb{R}^{9088}$.

    \subsection{Face Reenactment in the style space $\mathcal{S}$}\label{ssec:method_S}
        Given a pair of random latent codes in the $\mathcal{Z}$ space, one for the source and one for the target image, i.e., $\mathbf{z}_s,\mathbf{z}_t$ , we obtain the corresponding codes in $\mathcal{W}$ space using the StyleGAN2's mapping network. The resulting codes, $\mathbf{w}_s$ and $\mathbf{w}_t$, are used to calculate the style codes $\mathbf{s}_s$ and $\mathbf{s}_t$, using the affine module $A$, and ultimately generate the source ($I_s$) and the target ($I_t$) images, respectively, as shown in Fig.~\ref{fig:architecture}.
        Given the source and target style codes $\mathbf{s}_s$ and $\mathbf{s}_t$, we calculate a mask vector $\mathbf{m}$ such that the reenacted style code $\mathbf{s}_r\in\mathcal{S}$ contains the channels of $\mathbf{s}_t$ that correspond to the facial pose (i.e. head pose and expression) and the channels of $\mathbf{s}_s$ that correspond to the identity characteristics of a face (i.e. facial shape, hair color etc.). That is,
        \begin{equation}\label{eq:mask_eq}
            \mathbf{s}_r=\mathbf{m}\odot\mathbf{s}_t+(1-\mathbf{m})\odot\mathbf{s}_s,
        \end{equation}
        where $\odot$ denotes the element-wise multiplication. We do this by optimizing a \textit{Mask Network} on the difference of the two input style codes, i.e., on $\Delta\mathbf{s}=\mathbf{s}_s - \mathbf{s}_t$, as described below.
        
        %
        %
        \paragraph{\textbf{Mask Network}} As illustrated in Fig.~\ref{fig:architecture}, a style code $\mathbf{s}\in\mathcal{S}$ consists of $N_l$ style vectors $\mathbf{s}^i$, $i=1,\ldots,N_l$, each one corresponding to a different layer of the synthesis network $\mathcal{G}$. Following the basic hierarchical structure of StyleGAN~\cite{karras2019style}, where the first layers control coarse details (such as the head pose), the middle layers control semantic attributes (such as the expressions or the hair style), and the final layers control fine-grained details of the output images, we design our Mask Network in a similar manner. That is, we optimize a separate mask $\mathbf{m}_i$, $i=1,\ldots,N_l$, one for each input layer. Each sub-network of the Mask Network consists of two fully connected layers followed by a ReLU activation and the output of the last layer is passed through a sigmoid activation function. In order to obtain the final mask $\mathbf{m}$, we concatenate the output of each sub-network $\mathbf{m}_i$. In Section~\ref{ssec:ablation}, we show that this design choice leads to better results than using a single shared network for all layers.
        
        In order to train the Mask Network to disentangle the facial pose and identity characteristics we leverage the disentangled properties of the 3D Morphable Models (3DMMs)~\cite{blanz1999morphable}. Specifically, a 3D shape model $\textbf{X}\in\mathbb{R}^{3N}$, where $N$ is the number of facial landmark points, is defined as:
        \begin{equation}\label{eq:3dmm}
            X=\bar{X} + U_{s}\mathbf{a}_s + U_{e}\mathbf{a}_e,
        \end{equation}
        where $\bar{X}\in\mathbb{R}^{3N}$ denotes the mean 3D shape model, $U_s\in\mathbb{R}^{3N\times m_s}$ and $U_e\in\mathbb{R}^{3N\times m_e}$ denote the orthonormal bases of shape and expression, and $\mathbf{a}_s\in\mathbb{R}^{m_s}$ and $\mathbf{a}_e\in\mathbb{R}^{m_e}$ denote the corresponding shape and expression coefficients, respectively. We calculate the 3D shape models (i.e., $X_s$ and $X_t$) using the 3D reconstruction method proposed in~\cite{feng2021deca}. Moreover, using Eq.~\ref{eq:3dmm}, we are able to reconstruct the 3D shape model of the ground-truth reenacted face $X_t^{gt}$ using the source facial shape coefficients $\mathbf{a}_{s}$ and the target facial pose coefficients $\mathbf{a}_{e}$.
        
        %
        %
        \paragraph{Reenactment objective} We propose the minimization of the following loss term
        \begin{equation}\label{eq:reenact_loss}
            \mathcal{L}_r = \lambda_x\mathcal{L}_x + \lambda_{id}\mathcal{L}_{id},
        \end{equation}
        where $\lambda_x, \lambda_{id}$ are the loss coefficients and, $\mathcal{L}_x=\lVert X_r - X_t^{gt}\rVert_1$ denotes the shape loss and $\mathcal{L}_{id}$ the identity loss. Specifically, the shape loss ($\mathcal{L}_x$) is defined as the $L1$ distance between the reenacted 3D shape $X_r$ calculated using the reenacted image and the reconstructed ground-truth 3D shape $X_t^{gt}$. Moreover, following the common practice in neural face reenactment literature~\cite{bounareli2022finding,zakharov2020fast}, we use an identity preserving loss, between the source and the target images (since we do not use paired data during training), using ArcFace~\cite{deng2019arcface}. It is worth noting that imposing explicitly the preservation of the identity characteristics (via $\mathcal{L}_{id}$) is crucial, as shown in~\cite{bounareli2022finding}.
        
        %
        %
        \paragraph{Recurrent cycle consistency objective} In order to further enhance the identity preservation between the source and the reenacted images we use the recurrent cycle consistency loss~\cite{sanchez2020recurrent}. Specifically, we sample a new random target image $I_{t^\prime}$ and we use it to reenact both the source $I_s$ and the reenacted image $I_r$ as follows:
        
        \begin{equation}
          \begin{aligned}
         \mathbf{s}_r^1 = \mathbf{m}^1\odot\mathbf{s}_{t^\prime}+(1-\mathbf{m}^1)\odot\mathbf{s}_s,  \\
          \mathbf{s}_r^2 = \mathbf{m}^2\odot\mathbf{s}_{t^\prime}+(1-\mathbf{m}^2)\odot\mathbf{s}_r,
          \end{aligned}
        \end{equation}
        where the two reenacted images generated by the style codes $\mathbf{s}_r^1$ and $\mathbf{s}_r^2$ are encouraged to convey the source identity information and the facial pose of the second target image $I_{t^\prime}$. Consequently, the recurrent cycle consistency loss is defined as
        \begin{equation}\label{eq:cycle_loss}
            \mathcal{L}_{cycle} = \lambda_x\mathcal{L}_{cx}+\lambda_{id}\mathcal{L}_{cid},
        \end{equation}
        where the recurrent cycle shape loss, $\mathcal{L}_{cx}$, is given as
        \begin{equation}\label{eq:shape_cycle_loss}
            \mathcal{L}_{cx} = \lVert X_{r^1} - X_{t^\prime}^{gt}\rVert_1 + \lVert X_{r^2}-X_{t^\prime}^{gt}\lVert_1, 
        \end{equation}
        where $X_{r^1}$ and $X_{r^2}$ denote the 3D shapes calculated using the two reenacted images $I_r^1$ and $I_r^2$, and $X_{t^\prime}^{gt}$ denotes the ground-truth 3D shape calculated using the source facial shape coefficients $\mathbf{a}_s$ and the target facial pose coefficients $\mathbf{a}_e$ from the image $I_{t^\prime}$ (Eq.~\ref{eq:3dmm}). Finally, we define the recurrent cycle identity loss as
        \begin{align}
            \mathcal{L}_{cid} = & \operatorname{sim}\left(\mathcal{F}(I_s),\mathcal{F}(I_r^1)\right) + \operatorname{sim}\left(\mathcal{F}(I_s),\mathcal{F}(I_r^2)\right) + \\ \nonumber
            & \operatorname{sim}\left(\mathcal{F}(I_r^1),\mathcal{F}(I_r^2)\right),    
        \end{align}
        where $\operatorname{sim}(\cdot)$ denotes the cosine similarity caclulated on the ArcFace~\cite{deng2019arcface} features $\mathcal{F}(\cdot)$ between the source ($I_s$) and the two new reenacted images ($I_r^1$ and $I_r^2$), respectively.
        
        Finally, the total loss being minimized by the proposed framework during training is given as
        \begin{equation}\label{eq:loss_all}
            \mathcal{L} = \mathcal{L}_r + \mathcal{L}_{cycle},
        \end{equation}
        where $\mathcal{L}_{r}$ is the reenactment loss given by Eq.~\ref{eq:reenact_loss} and $\mathcal{L}_{cycle}$ is the recurrent cycle consistency loss given by Eq.~\ref{eq:cycle_loss}.

\section{Experiments}\label{sec:experiments}
    In this section, we briefly describe the implementation details for training the proposed method in Sect.~\ref{ssec:implementation_details}, we then discuss the adopted evaluation metrics in Sect.~\ref{ssec:evaluation_metrics}, and present qualitative and quantitative evaluation results in comparison to state-of-the-art methods for synthetic face reenactment ( Sect.~\ref{ssec:results}). Finally, in Sect.~\ref{ssec:ablation}, we present our ablation study on the various design choices of the proposed framework. We not that in the supplementary material we provide additional qualitative results for real face reenactment.
    
    \begin{table*}[t!]
            \centering
            \caption{Quantitative comparison of the proposed framework in comparison to the state-of-the-art Fast Bi-layer~\cite{zakharov2020fast}, ID-disentanglement (ID-d)~\cite{nitzan2020face}, PIR~\cite{ren2021pirenderer}, and StyleFusion~\cite{kafri2021stylefusion}.}
            \label{table:metrics}
            \begin{tabular}{|c||c|c|c|c|c|c|}
            \hline
            Method & CSIM $\uparrow$  & NME $\downarrow$ & Pose $\downarrow$  & Expr. $\downarrow$ & FID $\downarrow$ & Preference Rate (\%) $\uparrow$ \\ \hline
            Fast Bi-layer~\cite{zakharov2020fast} & 0.43 & 9.6 & 1.3 & 0.10 & 90.0 & 3.6\\
            ID-d~\cite{nitzan2020face} & 0.57 & 12.1 & 2.1 & 0.12 & 26.0 & 15.4\\
            PIR~\cite{ren2021pirenderer} & 0.69 & 11.1 & 1.6 & 0.11 & 27.4 & 26.0 \\
            StyleFusion~\cite{kafri2021stylefusion} & 0.63 & 18.1  & 3.75 & 0.12 & 11.0 & 9.0\\
            Ours & \textbf{0.70} & \textbf{9.2} & \textbf{1.2} & \textbf{0.09} & \textbf{10.0} & \textbf{46.0}\\ \hline
            \end{tabular}
        \end{table*}
    
    \subsection{Implementation details}\label{ssec:implementation_details}
        We use the StyleGAN2~\cite{karras2020analyzing} generator, trained on the FFHQ dataset~\cite{karras2019style}. The dimensionality of the style space $\mathcal{S}$ is $9088$, where $6048$ dimensions correspond to the convolutional layers and $3040$ to the ToRGB blocks of the generator $\mathcal{G}$. In our experiments we use the 12 first layers (out of the 18 layers) of the generator (i.e., $5632$ dimensions), since the rest affect only minor, fine-grained details of the image -- we note that using all layers leads to negligible boosts in the reenactment performance. Finally, we train our model for 70K iterations with a batch size of 6, Adam optimizer~\cite{kingma2014adam}, and learning rate $10^{-4}$.

    \begin{figure}[t]
        \centering
        \includegraphics[width=0.9\linewidth]{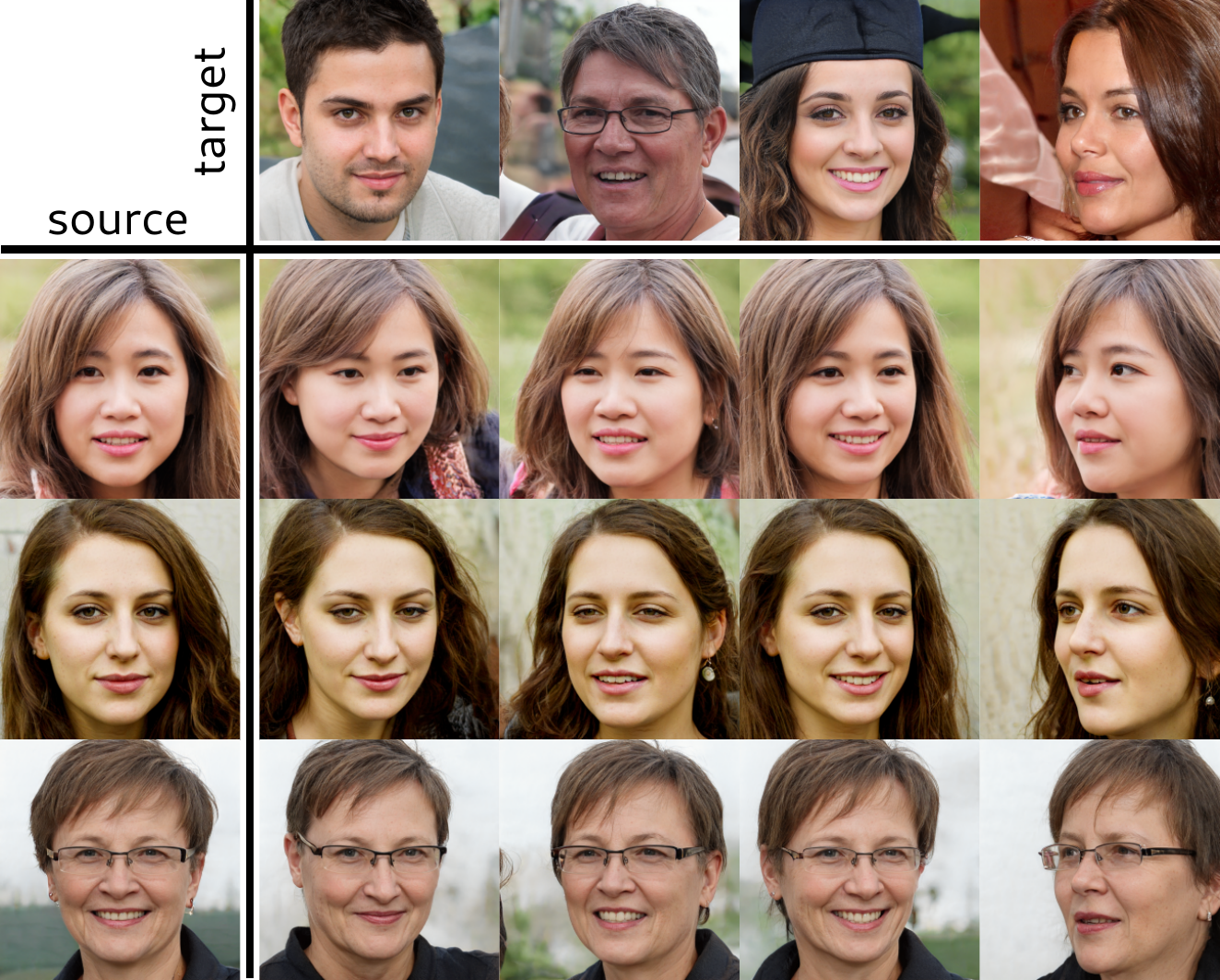}
        \caption{Qualitative results of the proposed method on face reenactment using synthetic images from StyleGAN2~\cite{karras2020analyzing}. Given two sets of three source and four target images, we show that our method arrives at reenacted faces where the source identity information (e.g., hair style or facial shape) is faithfully preserved, while head pose and facial expressions are effectively transferred to the target.}
        \label{fig:reenact_fig}
    \end{figure}
    
    \subsection{Evaluation metrics}\label{ssec:evaluation_metrics}
        For the quantitative evaluation of the proposed method and the state-of-the-art works we adopt five evaluation metrics, following the common practice in the related literature~\cite{bounareli2022finding,zakharov2020fast,ren2021pirenderer}. First, we report the ArcFace~\cite{deng2019arcface} identity similarity score to measure the identity preservation between the source and the reenacted images -- we denote this as ``CSIM''. Moreover, in order to measure the head pose transfer between the source and the target, we calculate the $L1$ distance of the differences for the three Euler angles (i.e., yaw, pitch, and roll) -- we denote this as ``Pose''. Similarly, in order to measure the facial expression transfer between the source and the target, we calculate the $L1$ distance of the differences for the expression coefficients $\mathbf{a}_{e}$ -- we denote this as ``Expr.''. Additionally, we report the normalized mean error~\cite{bulat2017far} -- denoted as ``NME'' -- using the 3D facial landmarks of the reenacted images and the ground-truth 3D facial landmarks extracted as described in Sect.~\ref{ssec:method_S} using Eq.~\ref{eq:3dmm}. Finally, we report the Frechet-Inception Distance (FID)~\cite{heusel2017gans} to measure the quality of the reenacted images. 
    
    \subsection{Face Reenactment}\label{ssec:results}
        In Fig.~\ref{fig:reenact_fig}, we show qualitative face reenactment results of the proposed method on synthetic images (using StyleGAN2~\cite{karras2020analyzing}). We show that the proposed method faithfully preserves the identity information of the source faces (e.g., in terms of hair style/colour, facial shape, etc), while at the same time effectively transfers the head pose (in terms of yaw, pitch, and roll) and the facial expressions of the target faces. Moreover, we note that additional accessories, such as eyeglasses (e.g., see source face in the third row) are preserved on the reenacted images, while accessories from target faces are not transferred onto the reenacted ones, such as the hat of the target face in the third column, or the eyeglasses of the target face in the second column. This clearly indicates that our model effectively disentangles the facial pose from the identity characteristics without transferring any style details from the target faces.
        
\begin{figure*}[t]
            \centering
            \includegraphics[width=0.7\textwidth]{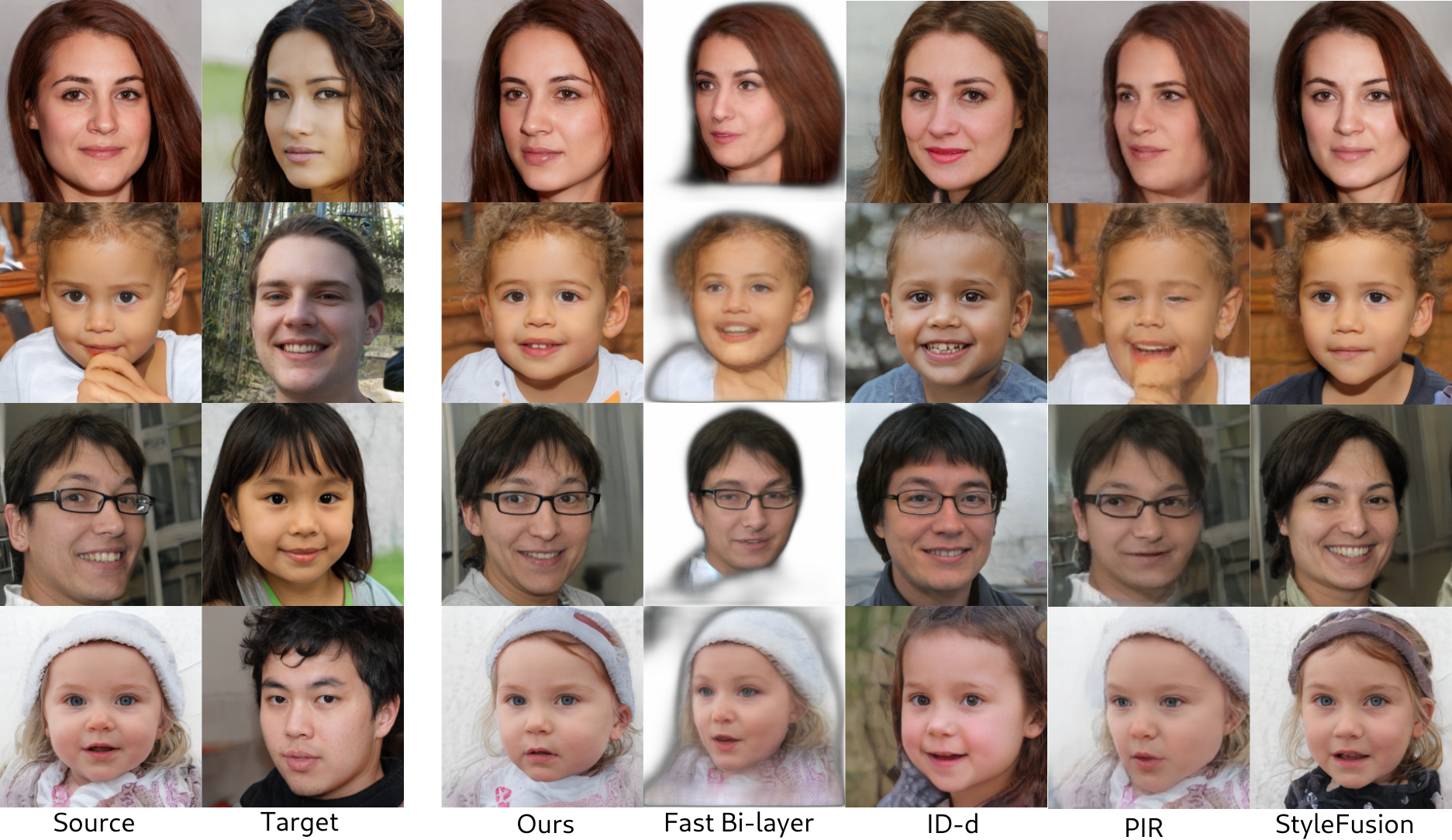}
            \caption{Qualitative comparison of the proposed method to the state-of-the-art Fast Bi-layer~\cite{zakharov2020fast}, ID-disentanglement (ID-d)~\cite{nitzan2020face}, PIR~\cite{ren2021pirenderer}, and StyleFusion~\cite{kafri2021stylefusion}.}
            \label{fig:comparison_fig}
        \end{figure*}

        We compare our proposed method to four state-of-the-art works, namely, Fast Bi-layer~\cite{zakharov2020fast}, ID-disentanglement (ID-d)~\cite{nitzan2020face}, StyleFusion~\cite{kafri2021stylefusion}, and PIR~\cite{ren2021pirenderer}. We note that in ID-disentanglement~\cite{nitzan2020face}, the authors trained an encoder-decoder architecture to disentangle the identity and the semantic attributes using a StyleGAN pretrained generator. In StyleFusion~\cite{kafri2021stylefusion}, the authors propose to fuse different style codes from synthetic images in order to generate an image composed by different regions of the input images. We use their model to fuse two input style codes by transferring only the head pose and the expression from the target style code. In order to compare with ID-disentanglement~\cite{nitzan2020face} and StyleFusion~\cite{kafri2021stylefusion}, we use the models provided by the authors. We also compare our method to Fast Bi-layer~\cite{zakharov2020fast} and PIR~\cite{ren2021pirenderer} using their models that are trained on the large-scale video datasets of VoxCeleb1~\cite{Nagrani17} and VoxCeleb2~\cite{Chung18b}. For our experiments and comparisons we randomly sample 5K image pairs, where the source and target faces have different identities. In Fig.~\ref{fig:comparison_fig}, we show qualitative results of the proposed method in comparison to the above four state-of-the-art works. We observe that our method arrives at much better reenactment performance, both in terms of identity preservation and head pose/expression transfer.
        
        In Table~\ref{table:metrics}, we report the quantitative evaluation of the proposed method in comparison to the state-of-the-art Fast Bi-layer~\cite{zakharov2020fast}, ID-disentanglement (ID-d)~\cite{nitzan2020face}, PIR~\cite{ren2021pirenderer}, and StyleFusion~\cite{kafri2021stylefusion}, on a set of 5K randomly generated image pairs. We note that our method outperforms both ID-disentanglement (ID-d)~\cite{nitzan2020face} and StyleFusion~\cite{kafri2021stylefusion} in all metrics. Regarding CSIM, our method achieves a score similar to PIR~\cite{ren2021pirenderer}, however, as shown in Fig.~\ref{fig:comparison_fig}, the reenacted images obtained by PIR~\cite{ren2021pirenderer} exhibit many visual artifacts that CSIM metric cannot capture. Moreover, we are able to better transfer the target facial pose. Also, we achieve competitive results against the state-of-the-art method of Fast Bi-layer~\cite{zakharov2020fast} which is trained on paired data with over 20K videos of multiple identities. We additionally compare our method against StyleFlow~\cite{abdal2021styleflow}, a state-of-the-art method that finds non-linear paths in the latent space of StyleGAN2. To compare with StyleFlow~\cite{abdal2021styleflow} we used a smaller set (500 image pairs) provided by the authors in~\cite{abdal2021styleflow}. As shown in Table~\ref{table:metrics_styleflow}, our method outperforms StyleFlow in all evaluation metrics. In the supplementary material, we provide additional qualitative comparisons to the above state-of-the-art works.
        
        Finally, we conducted a user study, where we show 20 randomly selected image pairs on $30$ users and asked them to select between the five methods (Fast Bi-layer~\cite{zakharov2020fast}, ID-disentanglement (ID-d)~\cite{nitzan2020face}, PIR~\cite{ren2021pirenderer}, StyleFusion~\cite{kafri2021stylefusion} and Ours) the one that best reenacts the source face. In Table~\ref{table:metrics}, we present the preference rate of users for each method. As shown, according to the user study, our method is the most preferable, which is consistent to the quantitative results.
        
        \begin{table}[t]
            \centering
            \caption{Quantitative comparison of the proposed method with StyleFlow~\cite{abdal2021styleflow}.}
            \label{table:metrics_styleflow}
            \begin{tabular}{|c||c|c|c|c|}
            \hline
            Method & CSIM $\uparrow$ & NME $\downarrow$ & Pose $\downarrow$  & Expr. $\downarrow$\\ \hline
            StyleFlow~\cite{abdal2021styleflow} & 0.67 & 15.8  & 2.6 & 0.13 \\
            Ours & \textbf{0.70} & \textbf{8.9} & \textbf{1.1} & \textbf{0.10} \\ \hline
            \end{tabular}
        \end{table}

    \subsection{Ablation Study}\label{ssec:ablation}
        In this section, we present our ablation study on various design choices of the proposed method. We note that for our ablation study we use $2K$ randomly generated image pairs and we report the evaluation metrics discussed above. First, we conduct an ablation study on the disentanglement properties of the adopted GAN latent space. Specifically, we compare the style space $\mathcal{S}$ with the standard $\mathcal{W}$ space and we report the results in Table~\ref{table:ablation_1}. Clearly, the style space $\mathcal{S}$ leads to better disentanglement, and thus better face reenactment, in terms of all the reported evaluation metrics.
        
        \begin{table}[t]
            \centering
            \caption{Ablation study on the disentanglement properties of $\mathcal{S}$ and $\mathcal{W}$ latent spaces.}
            \label{table:ablation_1}
            \begin{tabular}{|c||c|c|c|c|c|}
            \hline
            Method & CSIM $\uparrow$ & NME $\downarrow$  & Pose $\downarrow$  & Expr. $\downarrow$ \\
            \hline
            $\mathcal{S}$ space & \textbf{0.69} & \textbf{ 9.8} & \textbf{1.4} & \textbf{1.0}\\
            $\mathcal{W}$ space & 0.68 & 18.3 & 4.1 &  1.3 \\
            \hline
            \end{tabular}
        \end{table}
        
        Moreover, we conduct an ablation study on the architecture of the proposed Mask Network (Sect.~\ref{ssec:method_S}). Specifically, we compare the use of separate mask networks $M_i$, $i=1,\ldots,N_l$, one for each layer of the generator $\mathcal{G}$, against a single mask network $M$, global for all layers of the generator. We report the results in Table~\ref{table:ablation_2}, where we note that a separate mask network for each input layer of $G$ leads to better results. We attribute this to the hierarchical architecture of StyleGAN2 generator, which gradually generates images by controlling the input of each layer separately -- each layer of the generator controls different level of details on the generated images~\cite{karras2019style}. Similarly, our choice of using different mask networks for each layer allows for better disentangling the identity and facial attributes channels of style codes. 
        
        \begin{table}[t]
            \centering
            \caption{Ablation study on the architecture of Mask Network.}
            \label{table:ablation_2}
            \begin{tabular}{|c||c|c|c|c|c|}
            \hline
            Method & CSIM $\uparrow$ & NME $\downarrow$ & Pose $\downarrow$ & Expr. $\downarrow$ \\
            \hline
            Mask networks $M_i$  & \textbf{0.69} & \textbf{9.8} & \textbf{1.4} & \textbf{1.0} \\
            Global mask network $M$ &  0.64 & 10.7  & 1.6  & 1.1   \\
            \hline
            \end{tabular}
        \end{table}
        
        \begin{figure}[t]
            \centering
            \includegraphics[width=0.8\linewidth]{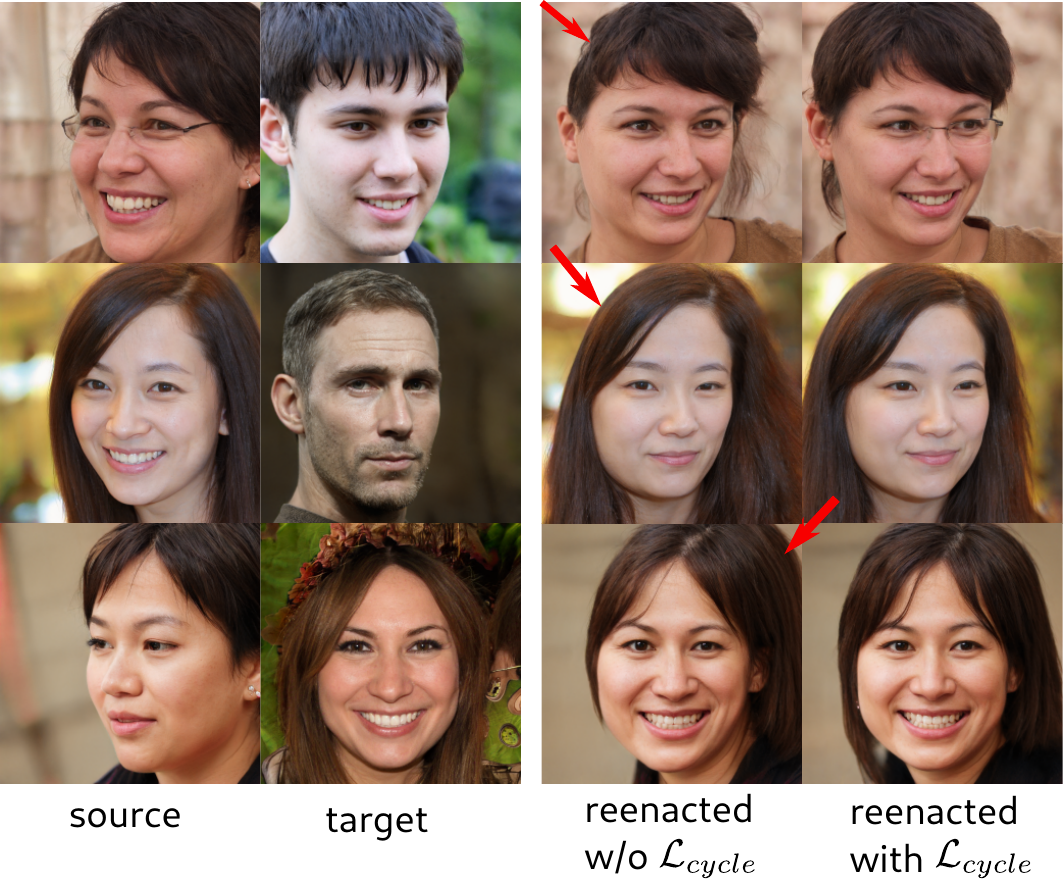}
            \caption{Ablation study on the use of the recurrent cycle loss $\mathcal{L}_{cycle}$. We observe that using the recurrent cycle loss reduces visual artifacts (marked with the red arrows).}
            \label{fig:ablation_fig}
        \end{figure}
        
        Finally, in order to assess the contribution of the recurrent cycle consistency loss $\mathcal{L}_{cycle}$, we report results of our method with and without using it in Table~\ref{table:ablation_3}. We note that using the additional recurrent cycle loss improves our results regarding the transfer of the facial pose as indicated by the metrics NME, Pose and Expr. However, the identity similarity (CSIM) is better without the cycle loss. In spite of this, we argue that the recurrent cycle consistency loss overall improves both our quantitative and qualitative results as shown in Fig.~\ref{fig:ablation_fig}. That is, our model, without the use of the recurrent cycle loss, produces results with some visual artifacts, especially in large pose variations between the source and target faces. Undoubtedly, the CSIM metric is not appropriate for capturing those artifacts. We also note that transferring an extreme target pose is challenging for most of the face reenactment methods. StyleFusion~\cite{kafri2021stylefusion} reports similar artifacts, especially on large pose variations between the source and target images. Nevertheless, incorporating the recurrent cycle consistency loss in the proposed method eliminates such artifacts, leading to better face reenactment. We additionally note that using the simple cycle loss described in~\cite{zhu2017unpaired} instead of the recurrent cycle loss~\cite{sanchez2020recurrent} does not improve the performance of our method.
        
        \begin{table}[t]
            \centering
            \caption{Ablation study regarding the use of the recurrent cycle consistency loss $\mathcal{L}_{cycle}$.}
            \label{table:ablation_3}
            \begin{tabular}{|c||c|c|c|c|c|}
            \hline
            Method & CSIM $\uparrow$ & NME $\downarrow$ & Pose $\downarrow$ & Expr. $\downarrow$\\
            \hline
            Ours with $\mathcal{L}_{cycle}$  & 0.69 & \textbf{9.4} & \textbf{1.1}  & \textbf{1.0} \\
            Ours w/o $\mathcal{L}_{cycle}$  & \textbf{0.72} & 9.9  & 1.5 & 1.2  \\
            \hline
            \end{tabular}
        \end{table}


\section{Conclusion}\label{sec:conclusion}
    
    In this paper we present a neural face reenactment method using the style space $\mathcal{S}$ of StyleGAN2. Specifically, given a source and a target style code, we learn to mask and mix them so that the reenacted style code consists of the channels of the target code that correspond to the facial pose and the channels of the source code that correspond to the identity characteristics. We show that our model can effectively transfer the target facial pose and preserve the source identity, even in the challenging case of extreme head poses, leading to state-of-the-art performance. Our model is trained on randomly generated samples, instead of paired training data, using supervision from a 3D shape model. On inference, the proposed framework relies solely on a pair of a source and a target image. Finally, the proposed method can straightforwardly be adapted for real image face reenactment, by incorporating a pre-trained GAN inversion method.

\bibliographystyle{IEEEtran}
\bibliography{main}

\begin{thebibliography}{10}
\providecommand{\url}[1]{#1}
\csname url@samestyle\endcsname
\providecommand{\newblock}{\relax}
\providecommand{\bibinfo}[2]{#2}
\providecommand{\BIBentrySTDinterwordspacing}{\spaceskip=0pt\relax}
\providecommand{\BIBentryALTinterwordstretchfactor}{4}
\providecommand{\BIBentryALTinterwordspacing}{\spaceskip=\fontdimen2\font plus
\BIBentryALTinterwordstretchfactor\fontdimen3\font minus
  \fontdimen4\font\relax}
\providecommand{\BIBforeignlanguage}[2]{{%
\expandafter\ifx\csname l@#1\endcsname\relax
\typeout{** WARNING: IEEEtran.bst: No hyphenation pattern has been}%
\typeout{** loaded for the language `#1'. Using the pattern for}%
\typeout{** the default language instead.}%
\else
\language=\csname l@#1\endcsname
\fi
#2}}
\providecommand{\BIBdecl}{\relax}
\BIBdecl

\bibitem{wu2021stylespace}
Z.~Wu, D.~Lischinski, and E.~Shechtman, ``Stylespace analysis: Disentangled
  controls for stylegan image generation,'' in \emph{Proceedings of the
  IEEE/CVF Conference on Computer Vision and Pattern Recognition}, 2021.

\bibitem{karras2020analyzing}
T.~Karras, S.~Laine, M.~Aittala, J.~Hellsten, J.~Lehtinen, and T.~Aila,
  ``Analyzing and improving the image quality of stylegan,'' in
  \emph{Proceedings of the IEEE/CVF Conference on Computer Vision and Pattern
  Recognition}, 2020.

\bibitem{goodfellow2014generative}
I.~Goodfellow, J.~Pouget-Abadie, M.~Mirza, B.~Xu, D.~Warde-Farley, S.~Ozair,
  A.~Courville, and Y.~Bengio, ``Generative adversarial nets,'' \emph{Advances
  in neural information processing systems}, vol.~27, 2014.

\bibitem{karras2019style}
T.~Karras, S.~Laine, and T.~Aila, ``A style-based generator architecture for
  generative adversarial networks,'' in \emph{Proceedings of the IEEE/CVF
  Conference on Computer Vision and Pattern Recognition}, 2019.

\bibitem{wang2018esrgan}
X.~Wang, K.~Yu, S.~Wu, J.~Gu, Y.~Liu, C.~Dong, Y.~Qiao, and C.~Change~Loy,
  ``Esrgan: Enhanced super-resolution generative adversarial networks,'' in
  \emph{Proceedings of the European conference on computer vision (ECCV)
  workshops}, 2018.

\bibitem{shen2020interfacegan}
Y.~Shen, C.~Yang, X.~Tang, and B.~Zhou, ``Interfacegan: Interpreting the
  disentangled face representation learned by gans,'' \emph{IEEE transactions
  on pattern analysis and machine intelligence}, 2020.

\bibitem{voynov2020unsupervised}
A.~Voynov and A.~Babenko, ``Unsupervised discovery of interpretable directions
  in the gan latent space,'' in \emph{International Conference on Machine
  Learning}.\hskip 1em plus 0.5em minus 0.4em\relax PMLR, 2020.

\bibitem{tzelepis2021warpedganspace}
C.~Tzelepis, G.~Tzimiropoulos, and I.~Patras, ``Warpedganspace: Finding
  non-linear rbf paths in gan latent space,'' in \emph{Proceedings of the
  IEEE/CVF International Conference on Computer Vision}, 2021.

\bibitem{oldfield2021tensor}
J.~Oldfield, M.~Georgopoulos, Y.~Panagakis, M.~A. Nicolaou, and I.~Patras,
  ``Tensor component analysis for interpreting the latent space of gans,''
  \emph{arXiv preprint arXiv:2111.11736}, 2021.

\bibitem{oldfield2022panda}
J.~Oldfield, C.~Tzelepis, Y.~Panagakis, M.~A. Nicolaou, and I.~Patras, ``Panda:
  Unsupervised learning of parts and appearances in the feature maps of gans,''
  \emph{arXiv preprint arXiv:2206.00048}, 2022.

\bibitem{tov2021designing}
O.~Tov, Y.~Alaluf, Y.~Nitzan, O.~Patashnik, and D.~Cohen-Or, ``Designing an
  encoder for stylegan image manipulation,'' \emph{ACM Transactions on Graphics
  (TOG)}, vol.~40, no.~4, pp. 1--14, 2021.

\bibitem{alaluf2021restyle}
Y.~Alaluf, O.~Patashnik, and D.~Cohen-Or, ``Restyle: A residual-based stylegan
  encoder via iterative refinement,'' in \emph{Proceedings of the IEEE/CVF
  International Conference on Computer Vision}, 2021.

\bibitem{alaluf2021hyperstyle}
Y.~Alaluf, O.~Tov, R.~Mokady, R.~Gal, and A.~H. Bermano, ``Hyperstyle: Stylegan
  inversion with hypernetworks for real image editing,'' \emph{arXiv preprint
  arXiv:2111.15666}, 2021.

\bibitem{zakharov2020fast}
E.~Zakharov, A.~Ivakhnenko, A.~Shysheya, and V.~Lempitsky, ``Fast bi-layer
  neural synthesis of one-shot realistic head avatars,'' in \emph{ECCV}, 2020.

\bibitem{bounareli2022finding}
S.~Bounareli, V.~Argyriou, and G.~Tzimiropoulos, ``Finding directions in gan's
  latent space for neural face reenactment,'' \emph{arXiv preprint
  arXiv:2202.00046}, 2022.

\bibitem{ren2021pirenderer}
Y.~Ren, G.~Li, Y.~Chen, T.~H. Li, and S.~Liu, ``Pirenderer: Controllable
  portrait image generation via semantic neural rendering,'' in
  \emph{Proceedings of the IEEE/CVF International Conference on Computer
  Vision}, 2021.

\bibitem{2020ganspace}
E.~Härkönen, A.~Hertzmann, J.~Lehtinen, and S.~Paris, ``Ganspace: Discovering
  interpretable gan controls,'' in \emph{Proc. NeurIPS}, 2020.

\bibitem{nitzan2020face}
Y.~Nitzan, A.~Bermano, Y.~Li, and D.~Cohen-Or, ``Face identity disentanglement
  via latent space mapping,'' \emph{arXiv preprint arXiv:2005.07728}, 2020.

\bibitem{doukas2020headgan}
M.~C. Doukas, S.~Zafeiriou, and V.~Sharmanska, ``Headgan: One-shot neural head
  synthesis and editing,'' in \emph{Proceedings of the IEEE/CVF International
  Conference on Computer Vision}, 2021.

\bibitem{siarohin2019first}
A.~Siarohin, S.~Lathuili{\`e}re, S.~Tulyakov, E.~Ricci, and N.~Sebe, ``First
  order motion model for image animation,'' \emph{Advances in Neural
  Information Processing Systems}, vol.~32, 2019.

\bibitem{Nagrani17}
A.~Nagrani, J.~S. Chung, and A.~Zisserman, ``Voxceleb: a large-scale speaker
  identification dataset,'' in \emph{INTERSPEECH}, 2017.

\bibitem{Chung18b}
J.~S. Chung, A.~Nagrani, and A.~Zisserman, ``Voxceleb2: Deep speaker
  recognition,'' in \emph{INTERSPEECH}, 2018.

\bibitem{zakharov2019few}
E.~Zakharov, A.~Shysheya, E.~Burkov, and V.~Lempitsky, ``Few-shot adversarial
  learning of realistic neural talking head models,'' in \emph{Proceedings of
  the IEEE/CVF International Conference on Computer Vision}, 2019.

\bibitem{kafri2021stylefusion}
O.~Kafri, O.~Patashnik, Y.~Alaluf, and D.~Cohen-Or, ``Stylefusion: A generative
  model for disentangling spatial segments,'' \emph{arXiv preprint
  arXiv:2107.07437}, 2021.

\bibitem{deng2019arcface}
J.~Deng, J.~Guo, N.~Xue, and S.~Zafeiriou, ``Arcface: Additive angular margin
  loss for deep face recognition,'' in \emph{Proceedings of the IEEE/CVF
  conference on computer vision and pattern recognition}, 2019.

\bibitem{feng2021deca}
Y.~Feng, H.~Feng, M.~J. Black, and T.~Bolkart, ``Learning an animatable
  detailed 3d face model from in-the-wild images,'' \emph{ACM Transactions on
  Graphics (TOG)}, vol.~40, no.~4, pp. 1--13, 2021.

\bibitem{abdal2021styleflow}
R.~Abdal, P.~Zhu, N.~J. Mitra, and P.~Wonka, ``Styleflow: Attribute-conditioned
  exploration of stylegan-generated images using conditional continuous
  normalizing flows,'' \emph{ACM Transactions on Graphics (ToG)}, vol.~40,
  no.~3, 2021.

\bibitem{tewari2020stylerig}
A.~Tewari, M.~Elgharib, G.~Bharaj, F.~Bernard, H.-P. Seidel, P.~P{\'e}rez,
  M.~Zollhofer, and C.~Theobalt, ``Stylerig: Rigging stylegan for 3d control
  over portrait images,'' in \emph{Proceedings of the IEEE/CVF Conference on
  Computer Vision and Pattern Recognition}, 2020.

\bibitem{shen2021closedform}
Y.~Shen and B.~Zhou, ``Closed-form factorization of latent semantics in gans,''
  in \emph{CVPR}, 2021.

\bibitem{patashnik2021styleclip}
O.~Patashnik, Z.~Wu, E.~Shechtman, D.~Cohen-Or, and D.~Lischinski, ``Styleclip:
  Text-driven manipulation of stylegan imagery,'' in \emph{Proceedings of the
  IEEE/CVF International Conference on Computer Vision}, 2021.

\bibitem{tzelepis2022contraclip}
C.~Tzelepis, J.~Oldfield, G.~Tzimiropoulos, and I.~Patras, ``Contraclip:
  Interpretable gan generation driven by pairs of contrasting sentences,''
  \emph{arXiv preprint arXiv:2206.02104}, 2022.

\bibitem{blanz1999morphable}
V.~Blanz and T.~Vetter, ``A morphable model for the synthesis of 3d faces,'' in
  \emph{Proceedings of the 26th annual conference on Computer graphics and
  interactive techniques}, 1999.

\bibitem{meshry2021learned}
M.~Meshry, S.~Suri, L.~S. Davis, and A.~Shrivastava, ``Learned spatial
  representations for few-shot talking-head synthesis,'' in \emph{Proceedings
  of the IEEE/CVF International Conference on Computer Vision}, 2021.

\bibitem{wang2021one}
T.-C. Wang, A.~Mallya, and M.-Y. Liu, ``One-shot free-view neural talking-head
  synthesis for video conferencing,'' in \emph{Proceedings of the IEEE/CVF
  Conference on Computer Vision and Pattern Recognition}, 2021.

\bibitem{abdal2019image2stylegan}
R.~Abdal, Y.~Qin, and P.~Wonka, ``Image2stylegan: How to embed images into the
  stylegan latent space?'' in \emph{Proceedings of the IEEE/CVF International
  Conference on Computer Vision}, 2019.

\bibitem{sanchez2020recurrent}
E.~Sanchez and M.~Valstar, ``A recurrent cycle consistency loss for progressive
  face-to-face synthesis,'' in \emph{2020 15th IEEE International Conference on
  Automatic Face and Gesture Recognition (FG 2020)}.\hskip 1em plus 0.5em minus
  0.4em\relax IEEE, 2020.

\bibitem{kingma2014adam}
D.~P. Kingma and J.~Ba, ``Adam: {A} method for stochastic optimization,'' in
  \emph{3rd International Conference on Learning Representations, {ICLR} 2015,
  San Diego, CA, USA, May 7-9, 2015, Conference Track Proceedings}, Y.~Bengio
  and Y.~LeCun, Eds., 2015.

\bibitem{bulat2017far}
A.~Bulat and G.~Tzimiropoulos, ``How far are we from solving the 2d \& 3d face
  alignment problem?(and a dataset of 230,000 3d facial landmarks),'' in
  \emph{Proceedings of the IEEE International Conference on Computer Vision},
  2017.

\bibitem{heusel2017gans}
M.~Heusel, H.~Ramsauer, T.~Unterthiner, B.~Nessler, and S.~Hochreiter, ``Gans
  trained by a two time-scale update rule converge to a local nash
  equilibrium,'' \emph{Advances in neural information processing systems},
  vol.~30, 2017.

\bibitem{zhu2017unpaired}
J.-Y. Zhu, T.~Park, P.~Isola, and A.~A. Efros, ``Unpaired image-to-image
  translation using cycle-consistent adversarial networks,'' in
  \emph{Proceedings of the IEEE international conference on computer vision},
  2017.

\end{thebibliography}

\clearpage
\appendices

\section{Supplementary material}

In this supplementary material, we will provide further details on the network architecture of the proposed framework in Sect.~\ref{sec:arch_details}, we will discuss its challenges and limitations in Sect.~\ref{sec:limitations}, and we will provide additional qualitative results on synthetic and real images, in comparison to state-of-the-art works, in Sect.~\ref{sec:results}.

\subsection{Network architecture details}\label{sec:arch_details}
\begin{figure}[!h]
        \centering
        \includegraphics[width=0.9\linewidth]{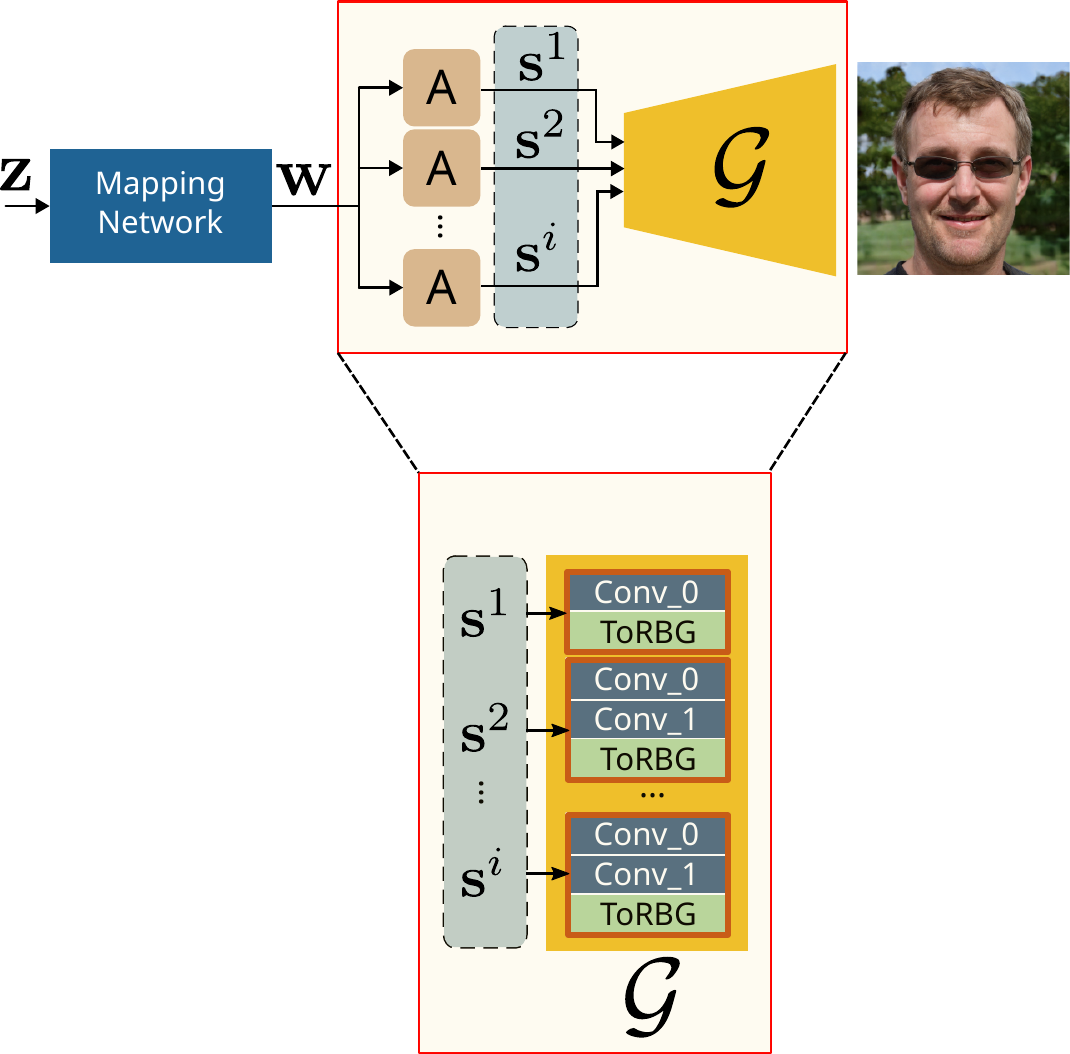}
        \caption{Hierarchical structure of the StyleGAN2~\cite{karras2020analyzing} generator.}
        \label{fig:architecture_details_0}
    \end{figure}

    In this section, we will discuss in detail the architecture of the proposed framework. Specifically, the design of the proposed Mask Network (see Sect.~III-B in the main paper), draws inspiration by the hierarchical structure of the StyleGAN2~\cite{karras2020analyzing} generator. As shown in Fig.~\ref{fig:architecture_details_0}, the latent codes $\mathbf{w}\in\mathcal{W}_+$ are processed by affine transformations $A$ generating the channel-wise style vectors $\mathbf{s}\in\mathcal{S}$, which are passed into the different layers of the generator $\mathcal{G}$. In Table~\ref{tab:stylespace}, we show the structure of StyleGAN2 synthesis network as described in~\cite{wu2021stylespace}.
    
    In the proposed method, we only use the style codes that are inputs to the convolutional layers (conv) and we apply our mask on the first $n=12$ layers of a total of $N_l=18$ layers of the generator $\mathcal{G}$. Specifically, we use the following style codes $\mathbf{s}^0,\mathbf{s}^2,\mathbf{s}^3,\mathbf{s}^5,\mathbf{s}^6,\mathbf{s}^8,\mathbf{s}^9,\mathbf{s}^{11},\mathbf{s}^{12},\mathbf{s}^{14},\mathbf{s}^{15}$, and $\mathbf{s}^{17}$ (see Table~\ref{tab:stylespace}) -- thus, the total number of style channels we use is $5632$.
    
\begin{figure}[!t]
        \centering
        \includegraphics[width=\linewidth]{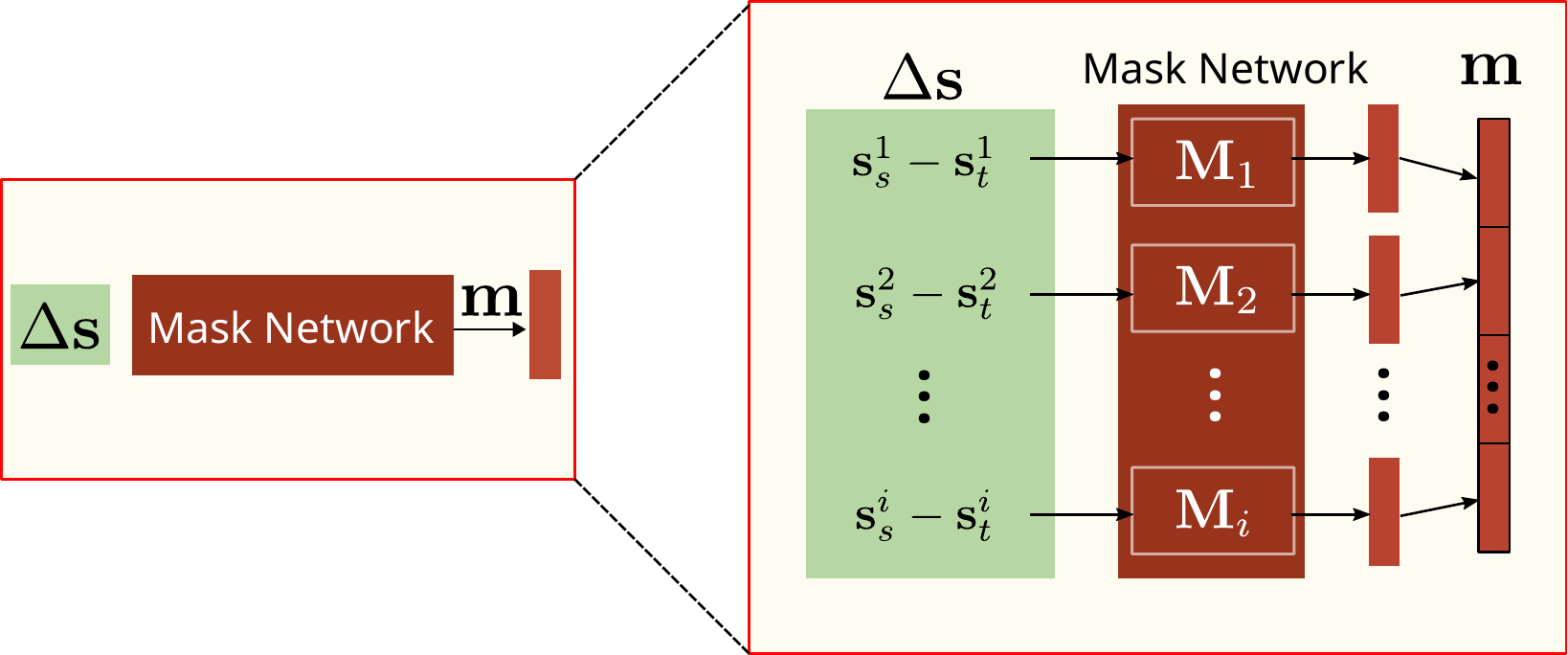}
        \caption{Architecture of the proposed Mask Network.}
        \label{fig:architecture_details_1}
    \end{figure}

    Finally, in Fig.~\ref{fig:architecture_details_1}, we illustrate in more detail the structure of the proposed Mask Network and its mask sub-networks $\mathbf{M}_i$, $i=1,\ldots,n$, where, as discussed above $n=12$. Specifically, we calculate the differences between the source and the target style codes, i.e., $\Delta\mathbf{s}_i=\mathbf{s}_s^i-\mathbf{s}_t^i$, $i=1,\ldots,n$, each of which is input to the corresponding mask sub-network $M_i$. Ultimately, the output vectors from all mask sub-networks are concatenated into the final mask vector $\mathbf{m}$. 
    
    \begin{table}[t]
        \centering
        \begin{tabular}{ccclc}
        $\mathcal{W+}$ index & $\mathcal{S}$ index & resolution & layer name & channels \\
        \rowcolor{Gray}
        0 & 0 & 4$\times$4 & Conv &  512 \\
        \rowcolor{Gray}
        1 & 1 & 4$\times$4 & ToRGB &  512 \\
        2 & 2 & 8$\times$8 & Conv0\_up & 512 \\
        3 & 3 & 8$\times$8 & Conv1 &  512 \\
        3 & 4 & 8$\times$8 & ToRGB &  512 \\
        \rowcolor{Gray}
        4 & 5 & 16$\times$16 & Conv0\_up &  512 \\
        \rowcolor{Gray}
        5 & 6 & 16$\times$16 & Conv1 & 512 \\
        \rowcolor{Gray}
        5 & 7 & 16$\times$16 & ToRGB &  512 \\
        6 & 8 & 32$\times$32 & Conv0\_up & 512 \\
        7 & 9 & 32$\times$32 & Conv1 & 512 \\
        7 & 10 & 32$\times$32 & ToRGB &  512 \\
        \rowcolor{Gray}
        8 & 11 & 64$\times$64 & Conv0\_up &  512 \\
        \rowcolor{Gray}
        9 & 12 & 64$\times$64 & Conv1 & 512 \\
        \rowcolor{Gray}
        9 & 13 & 64$\times$64 & ToRGB & 512 \\
        10 & 14 & 128$\times$128 & Conv0\_up &  512 \\
        11 & 15 & 128$\times$128 & Conv1 &  256 \\
        11 & 16 & 128$\times$128 & ToRGB &  256 \\
        \rowcolor{Gray}
        12 & 17 & 256$\times$256 & Conv0\_up & 256 \\
        \rowcolor{Gray}
        13 & 18 & 256$\times$256 & Conv1 & 128 \\
        \rowcolor{Gray}
        13 & 19 & 256$\times$256 & ToRGB & 128 \\
        14 & 20 & 512$\times$512 & Conv0\_up &  128 \\
        15 & 21 & 512$\times$512 & Conv1 &  64 \\
        15 & 22 & 512$\times$512 & ToRGB &  64 \\
        \rowcolor{Gray}
        16 & 23 & 1024$\times$1024 & Conv0\_up &  64 \\
        \rowcolor{Gray}
        17 & 24 & 1024$\times$1024 & Conv1 &  32 \\
        \rowcolor{Gray}
        17 & 25 & 1024$\times$1024 & ToRGB &  32
        \end{tabular}
        \caption{Structure of StyleGAN2 generator~\cite{wu2021stylespace}.} 
        \label{tab:stylespace}
    \end{table}


\subsection{Challenges and limitations}\label{sec:limitations}
    
\begin{figure}
        \centering
        \includegraphics[width=0.8\linewidth]{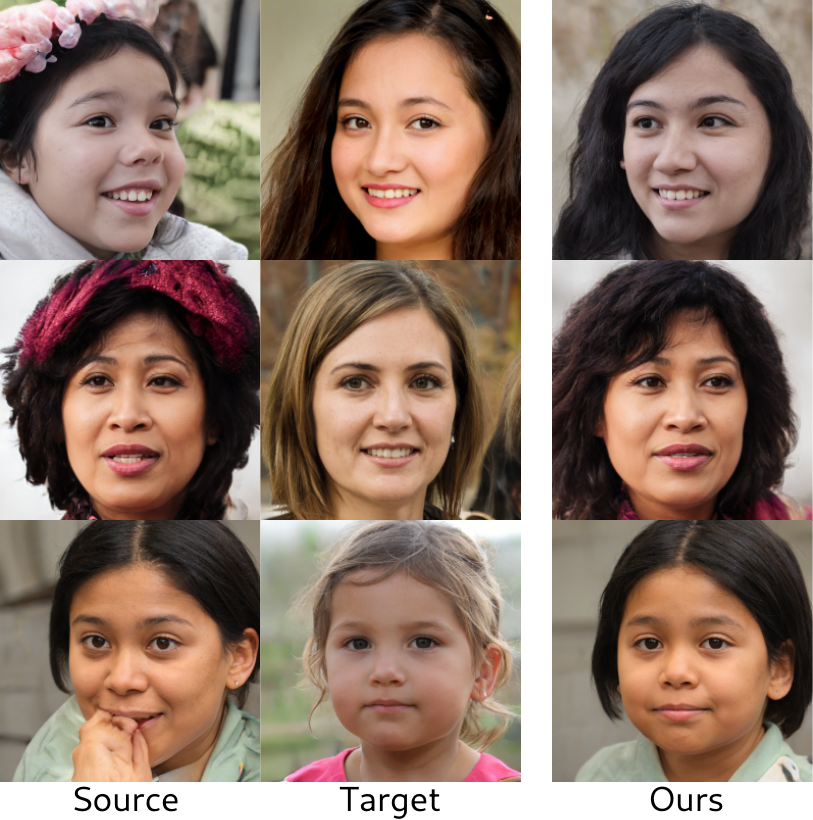}
        \caption{Arbitrary accessories or artifacts on source images are not transferred on the reenacted images.}
        \label{fig:limitations_0}
    \end{figure}

\begin{figure}
        \centering
        \includegraphics[width=0.8\linewidth]{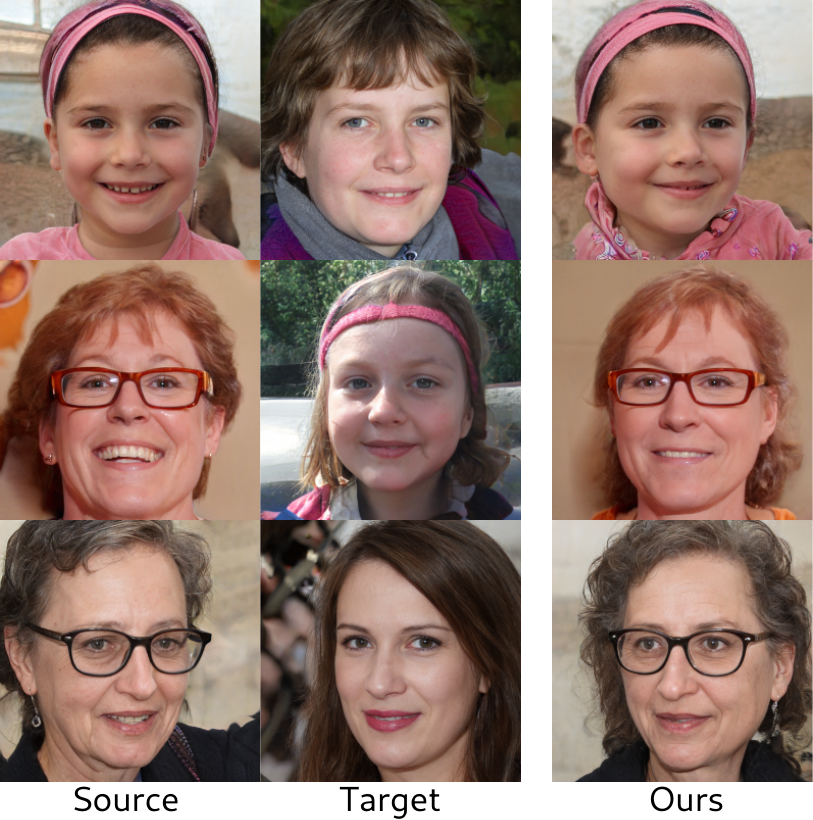}
        \caption{Well defined accessories such as hair accessories and glasses are preserved on the reenacted images.}
        \label{fig:limitations_0_compare}
    \end{figure}

    In this section we will discuss the challenges and the limitations of the proposed framework. We observe that when the source images that are generated by StyleGAN2 depict non-meaningful attributes or artifacts, these are not transferred on the reenacted images. For instance, as shown in Fig.~\ref{fig:limitations_0}, the two hair accessories on the first two source images and the noisy part around the mouth area of the third source image are not transferred on the reenacted images. In comparison, visually meaningful attributes, such as the hair accessory of the first source image in Fig.~\ref{fig:limitations_0_compare} or the eyeglasses of the second and the third source images in Fig.~\ref{fig:limitations_0_compare} are effectively transferred on the reenacted images. 
    
\begin{figure*}
        \centering
        \includegraphics[width=1.0\textwidth]{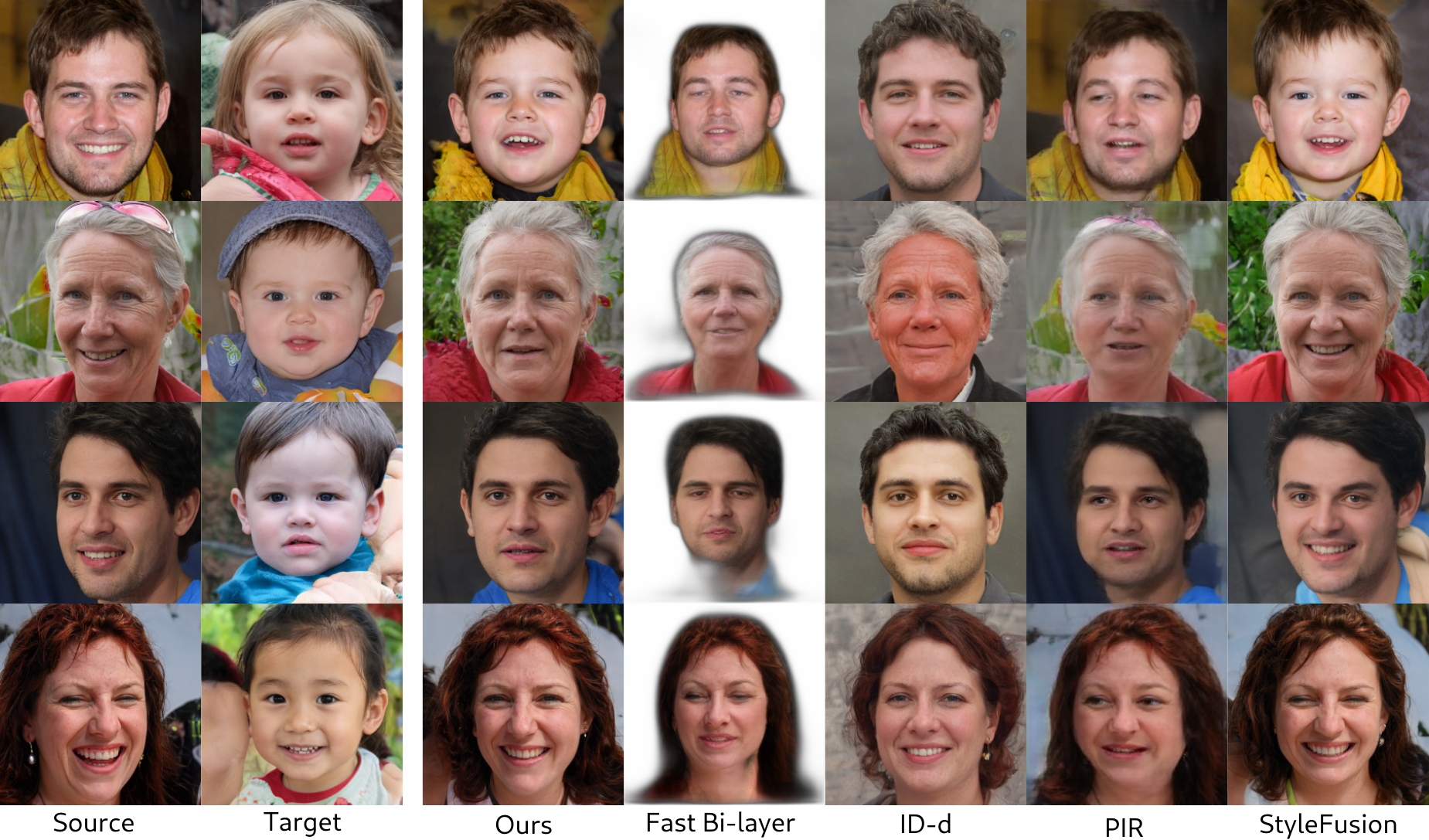}
        \caption{Face reenactment when the source and target faces have large facial shape differences. In the case of transferring the facial pose of younger target faces to older ones, most methods are not able to faithfully maintain the source facial shape. On the contrary, our method, except the first image pair, is able to maintain the source facial shape.}
        \label{fig:limitations_shape}
    \end{figure*}

    One of the main challenges of neural face reenactment concerns the effective transfer of the head pose and the expression of a target face onto a source face, without altering the source identity, even when the target and source faces have different identities. In small variations between the source and target faces, most methods provide compelling results, however, in case of large variations on the head pose or on the facial shape, most methods are not able to effectively reenact the source face. For instance, in Fig.~\ref{fig:limitations_shape}, we present some results on the challenging task of transferring the facial pose from a younger target face to an older one. It is worth noting that the facial shape of the source and target images are different. Most methods under such large facial shape differences are not able to faithfully preserve the source facial shape. By contrast, our method (except for the example pair on the first row of Fig.~\ref{fig:limitations_shape}), is able to correctly reenact the source faces, without transferring many identity characteristics from the target faces. 

\begin{figure*}
        \centering
        \includegraphics[width=1.0\textwidth]{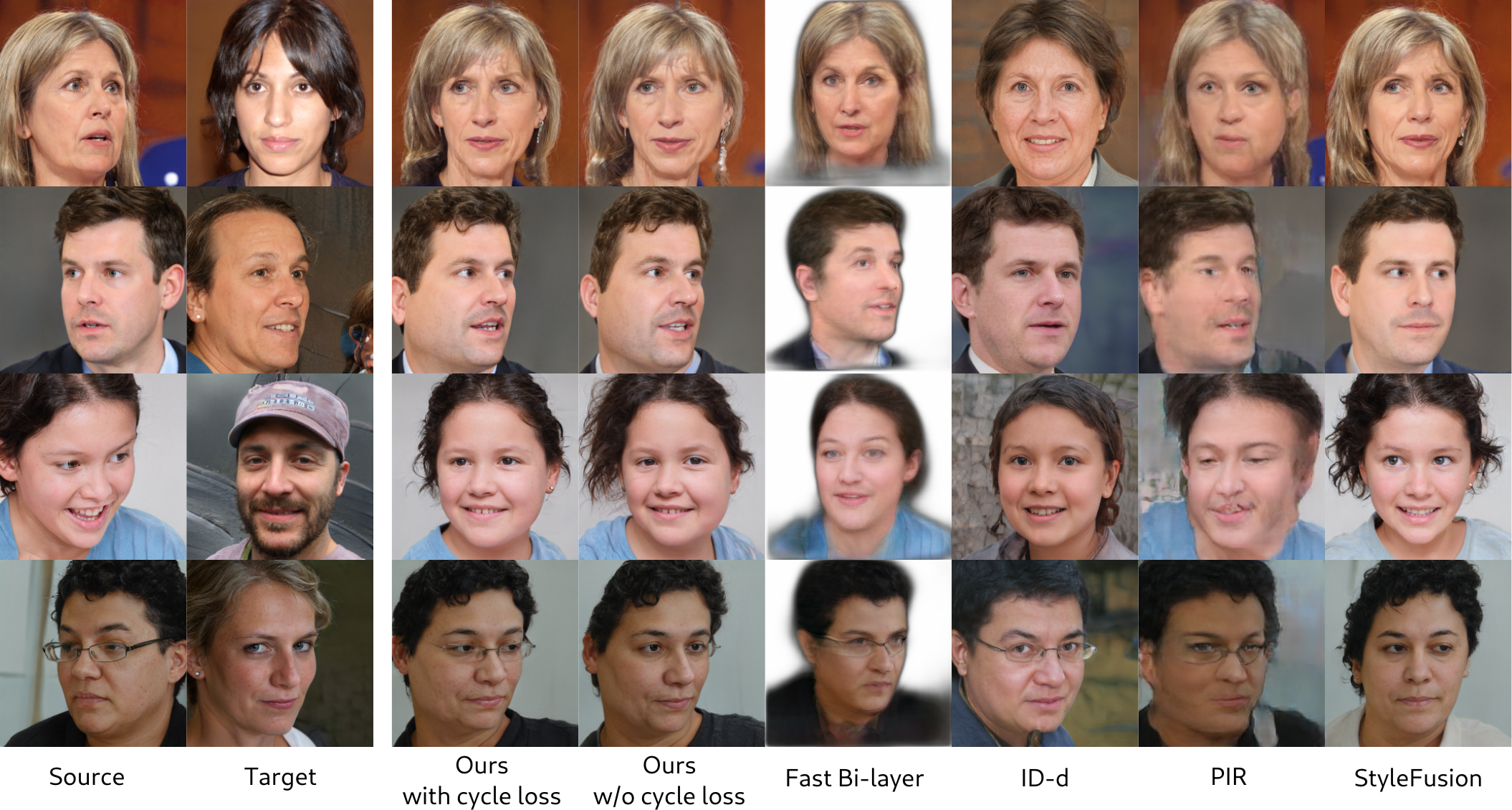}
        \caption{Face reenactment when the source and target faces have large pose variations. In the case of transferring the facial pose of a target face to a source face, when the head poses have large differences, most methods are not able to faithfully reenact the source images.}
        \label{fig:limitations_pose}
    \end{figure*}

    Finally, in Fig.~\ref{fig:limitations_pose}, we present some examples where the facial poses of the source and target faces have large differences. We observe that the reenacted images from the state-of-the-art Fast Bi-layer~\cite{zakharov2020fast}, PIR~\cite{ren2021pirenderer}, and StyleFusion~\cite{kafri2021stylefusion} introduce many visual artifacts, while StyleFusion~\cite{kafri2021stylefusion} is also not able to correctly transfer the target pose. Finally, despite the fact that ID-disentanglement (ID-d)~\cite{nitzan2020face} effectively transfers the pose, identity characteristics of the source faces are not well preserved.

\subsection{Additional qualitative results}\label{sec:results}

    \subsubsection{Face reenactment on synthetic images}
\begin{figure*}
            \centering
            \includegraphics[width=1.0\textwidth]{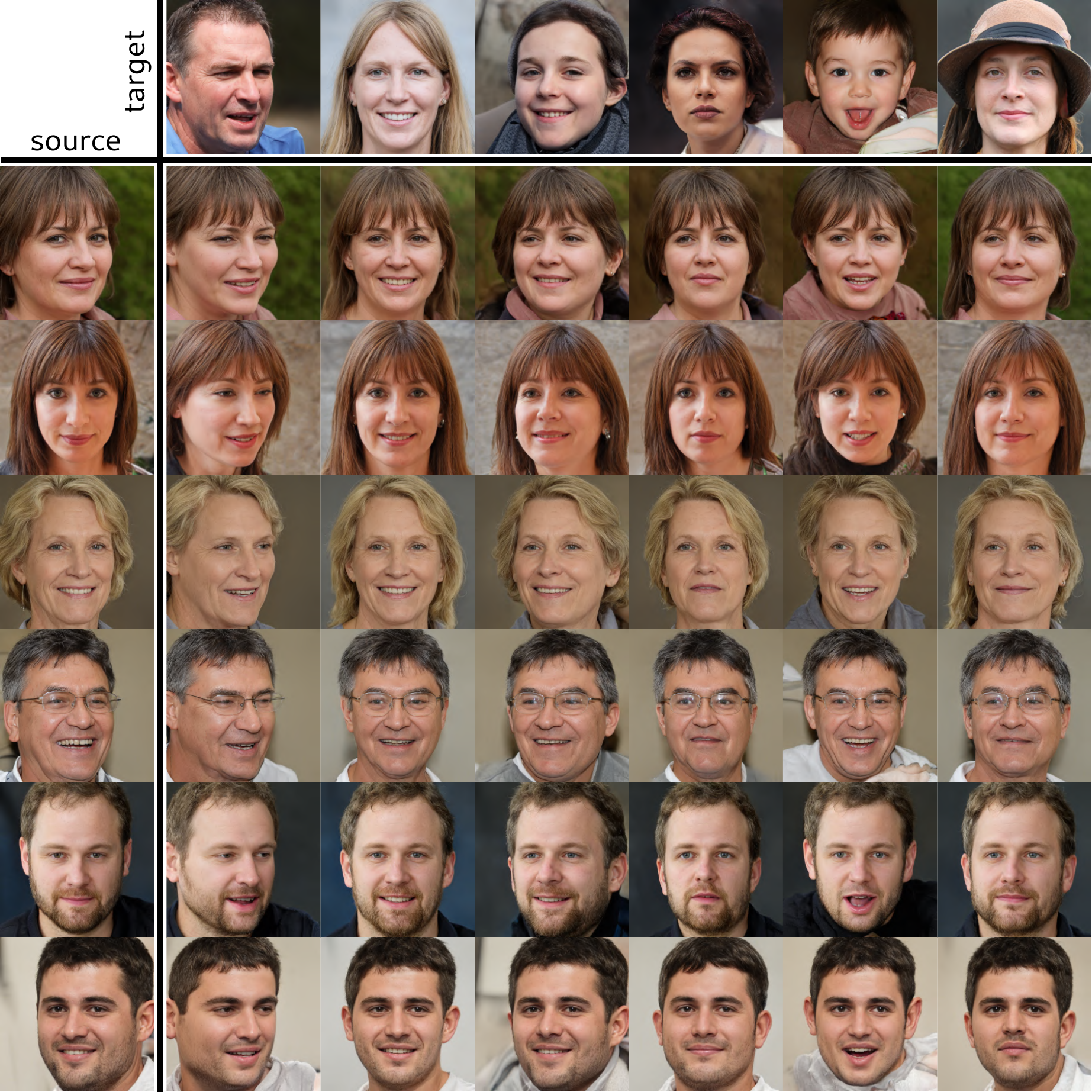}
            \caption{Additional qualitative results of our method on face reenactment using synthetic images.}
            \label{fig:source_target_fig}
        \end{figure*}

\begin{figure*}
            \centering
            \includegraphics[width=1.0\textwidth]{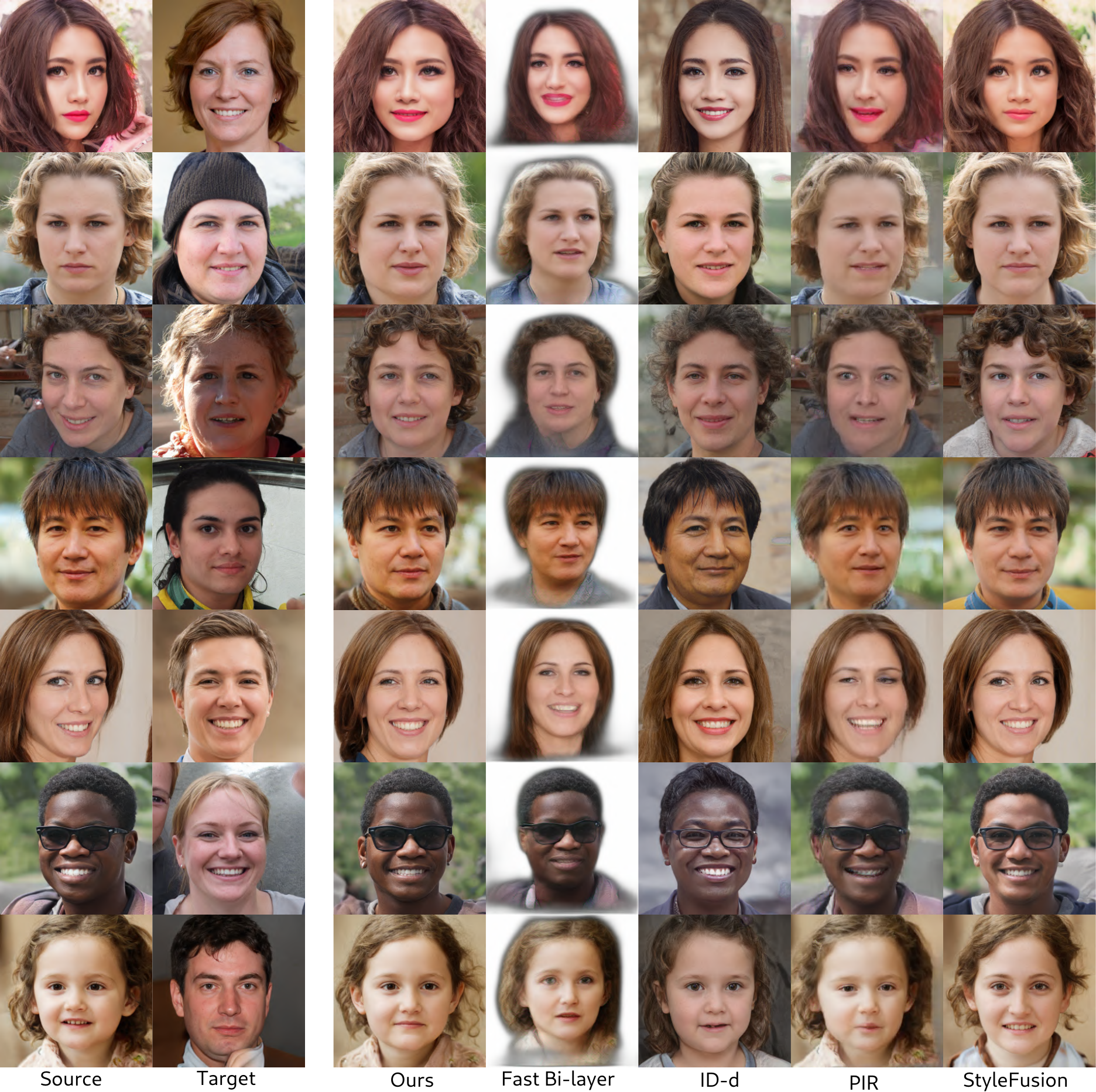}
            \caption{Qualitative comparison of the proposed method (``Ours'') with Fast Bi-layer~\cite{zakharov2020fast}, ID-disentanglement (ID-d)~\cite{nitzan2020face}, PIR~\cite{ren2021pirenderer}, and StyleFusion~\cite{kafri2021stylefusion}.}
            \label{fig:comparisons}
        \end{figure*}

\begin{figure*}
            \centering
            \includegraphics[width=0.7\textwidth]{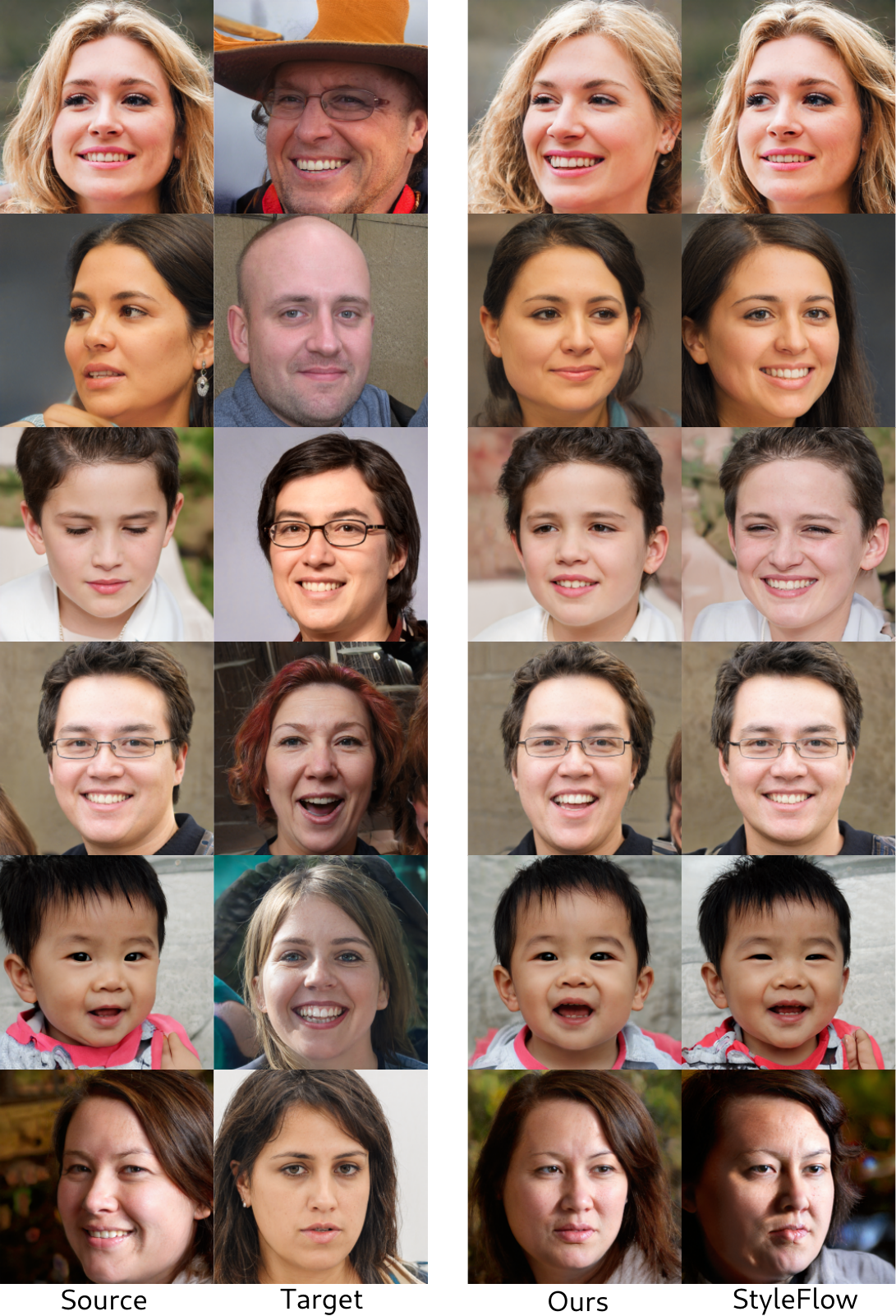}
            \caption{Qualitative comparison of the proposed method (``Ours'') with StyleFlow~\cite{abdal2021styleflow}.}
            \label{fig:comparison_styleflow}
        \end{figure*}

        In Fig.~\ref{fig:source_target_fig}, we show additional qualitative results of our method on face reenactment using randomly generated image pairs from StyleGAN2. Moreover, in Fig.~\ref{fig:comparisons} we show comparisons of our method against the state-of-the-art Fast Bi-layer~\cite{zakharov2020fast}, ID-disentanglement (ID-d)~\cite{nitzan2020face}, PIR~\cite{ren2021pirenderer}, and StyleFusion~\cite{kafri2021stylefusion}. Finally, in Fig.~\ref{fig:comparison_styleflow} we provide qualitative comparisons with StyleFlow~\cite{abdal2021styleflow} using randomly selected images from the image set provided by the authors of~\cite{abdal2021styleflow}. We observe that despite the fact that StyleFlow produces images without visual artifacts, it does not effectively transfer the target expression (e.g., image pair in the second row). Moreover, in large pose variations (e.g., image pairs in the third and the last rows), StyleFlow cannot preserve the identity characteristics.

    \subsubsection{Face reenactment on real images}
    
\begin{figure*}
            \centering
            \includegraphics[width=0.8\textwidth]{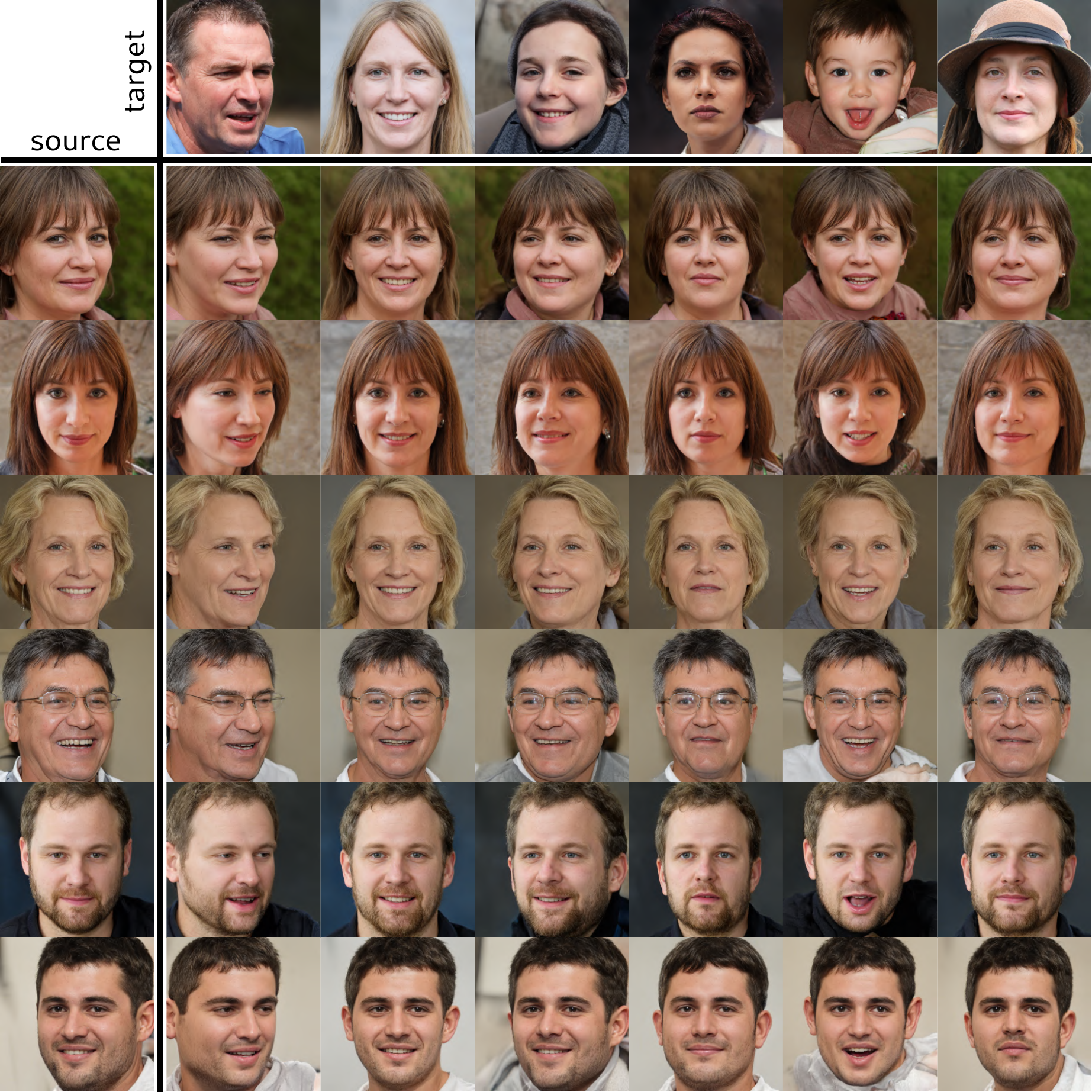}
            \caption{Qualitative results of our method on real images. Given a set of source and target images, we show results of reenacted images that preserve the source identity and have the target head pose and expression.}
            \label{fig:reenact_fig_real}
        \end{figure*}

        In this section, we present results of the proposed method in real face reenactment. To do so, we first embed the real faces into the latent space of the pre-trained StyleGAN2 using~\cite{tov2021designing}. In~\cite{tov2021designing}, Tov et al. propose an encoder network that is trained to predict a latent code in the $\mathcal{W}+$ space that best reconstructs the real image. Having the inverted latent codes, we calculate the corresponding style codes and then use our method to edit the real images. In Fig.~\ref{fig:reenact_fig_real} we show results of our method on pairs of real images from CelebA-HQ. 

\begin{figure*}
            \centering
            \includegraphics[width=1.0\textwidth]{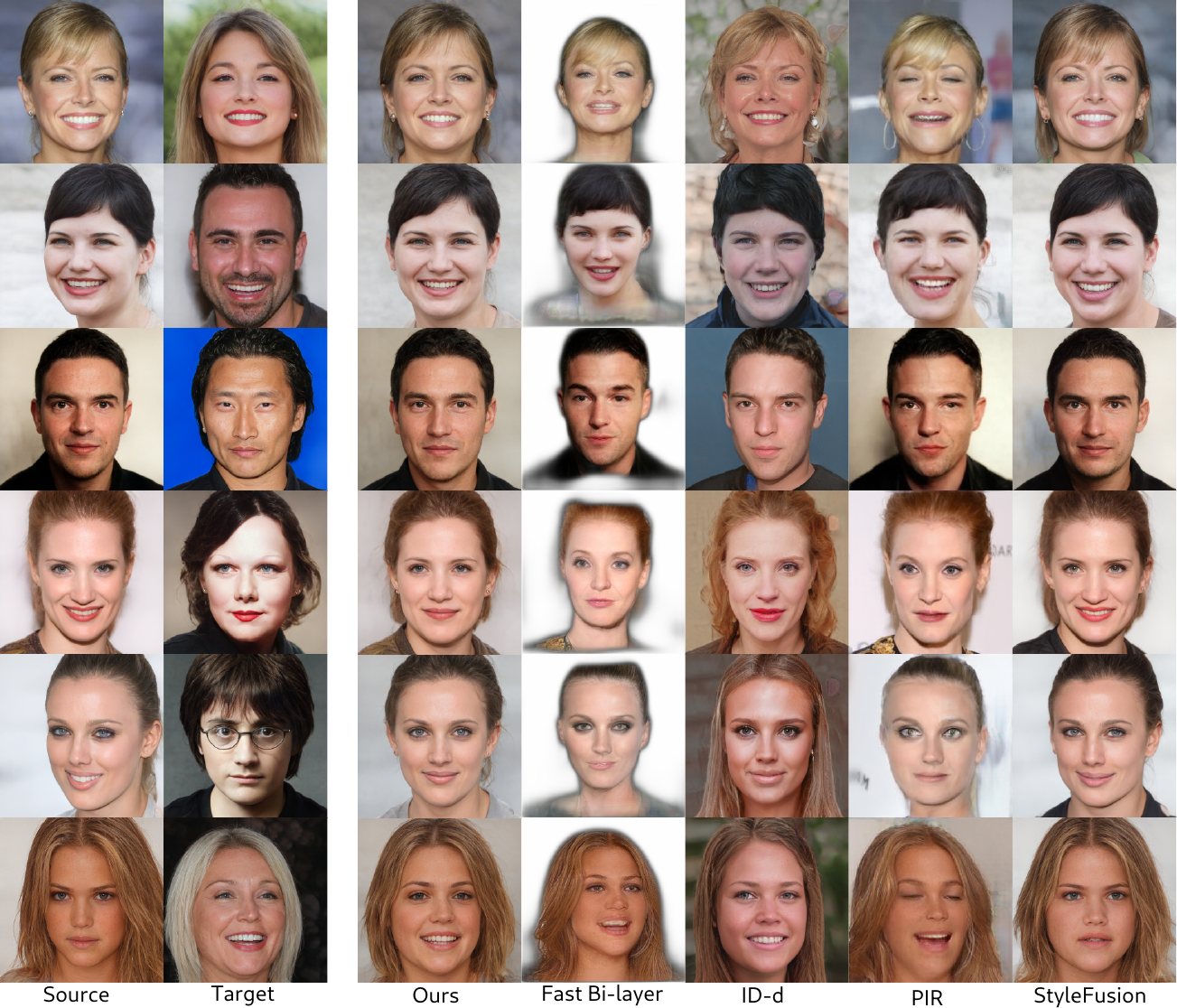}
            \caption{Qualitative comparison the proposed method (``Ours'') with Fast Bi-layer~\cite{zakharov2020fast}, ID-disentanglement (ID-d)~\cite{nitzan2020face}, PIR~\cite{ren2021pirenderer}, and StyleFusion~\cite{kafri2021stylefusion} on real images.}
            \label{fig:comparison_real}
        \end{figure*}

\begin{figure*}[ht]
            \begin{subfigure}{1.0\textwidth}
                \centering
                \includegraphics[width=0.9\linewidth]{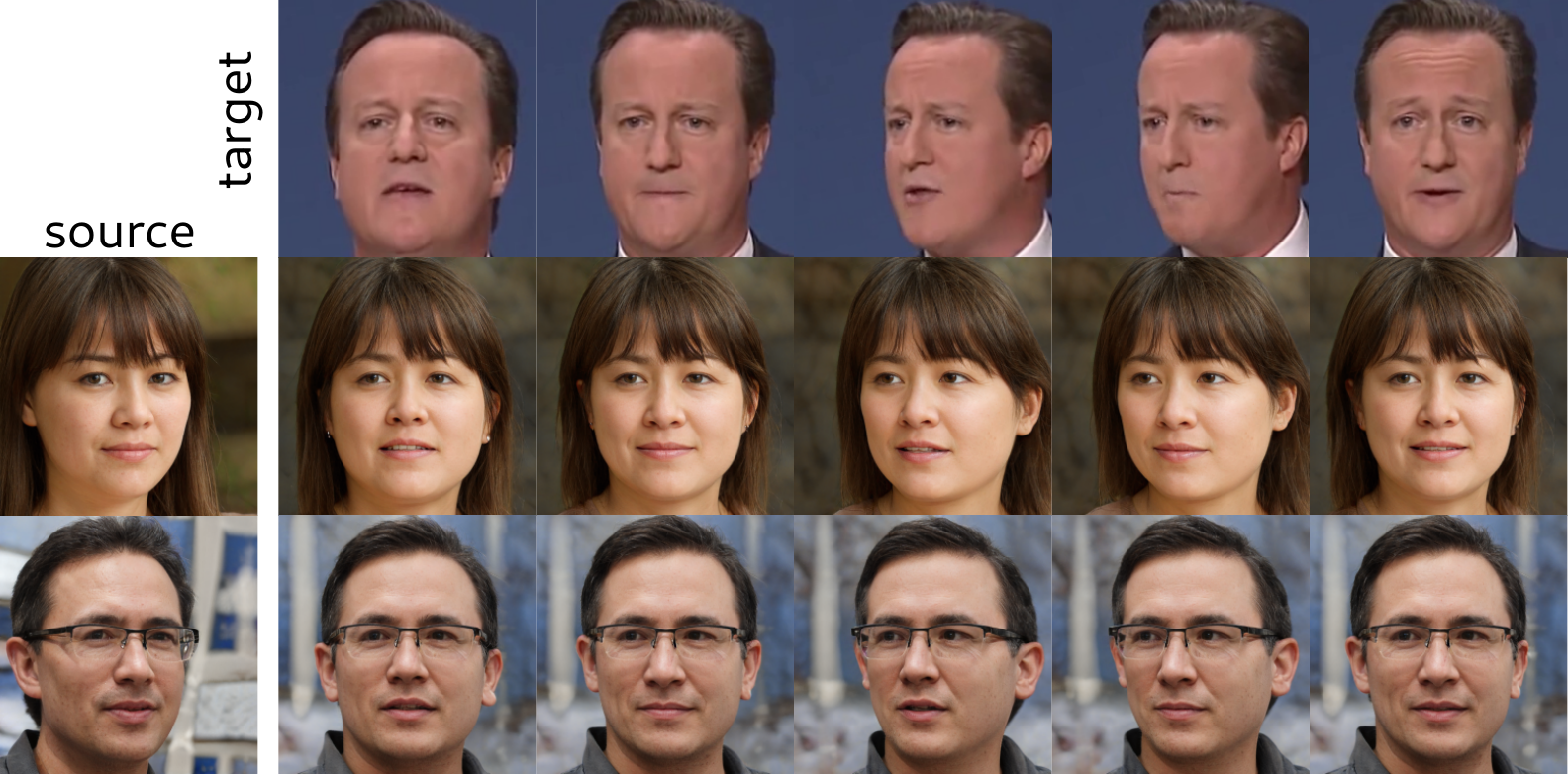}  
                \caption{Face reenactment of source synthetic images using as target video frames.}
                \label{fig:real_video_syn}
            \end{subfigure}
            \hfill
            \begin{subfigure}{1.0\textwidth}
                \centering
                \includegraphics[width=0.9\linewidth]{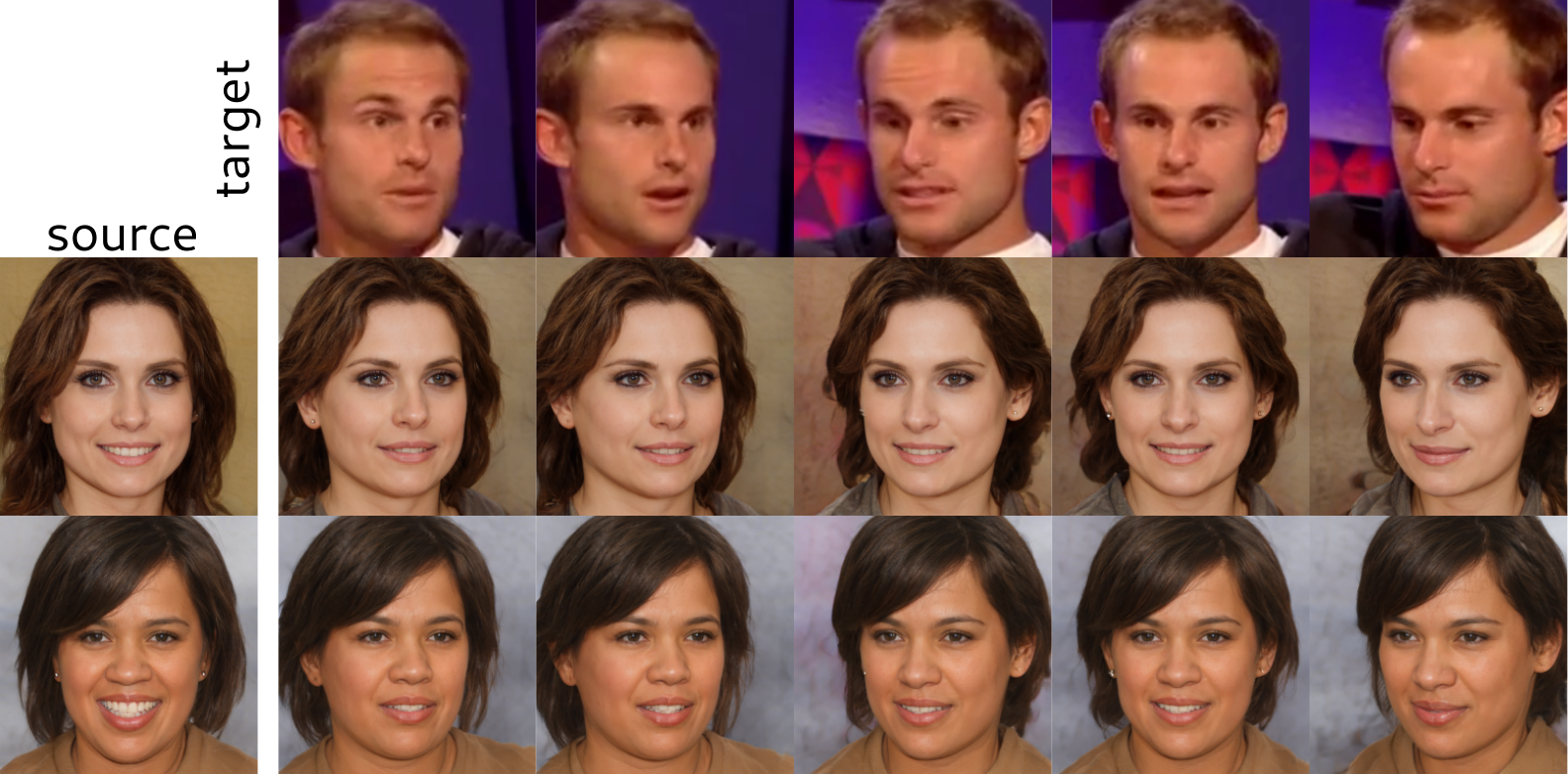}  
                \caption{Face reenactment of source real images using as target video frames.}
                \label{fig:real_video_real}
            \end{subfigure}
            \caption{Face reenactment of (a) synthetic and (b) real images using target faces from real video sequences.}
            \label{fig:real_video}
        \end{figure*}

        In Fig.~\ref{fig:comparison_real} we provide additional comparisons between our method and the state-of-the-art Fast Bi-layer~\cite{zakharov2020fast}, ID-disentanglement (ID-d)~\cite{nitzan2020face}, PIR~\cite{ren2021pirenderer}, and StyleFusion~\cite{kafri2021stylefusion} on real images from the CelebA-HQ dataset. Moreover, we show additional results on reenactment on synthetic and real images using as target video sequences. Specifically, as shown in Fig.~\ref{fig:real_video}, having multiple target frames of the same person, our method can effectively reenact a source face, either synthetic (Fig.~\ref{fig:real_video_syn}) or real (Fig.~\ref{fig:real_video_real}), in different facial poses and successfully preserve the source identity. We note that although the pretrained generator can synthesize high quality realistic images, it has a limitation on the variety of human expressions imposed by the FFHQ dataset. As a result, when applying our method on real images, where the range of facial attributes is different from the range in FFHQ dataset, there is a constraint on the expressions that our model can reconstruct.
    
\end{document}